\documentclass[10pt,twocolumn,letterpaper]{article}

\usepackage[pagenumbers]{cvpr} % To force page numbers, e.g. for an arXiv version

\definecolor{cvprblue}{rgb}{0.21,0.49,0.74}
\usepackage[pagebackref,breaklinks,colorlinks,allcolors=cvprblue]{hyperref}

\title{Img-Diff: Contrastive Data Synthesis for Multimodal Large Language Models}

\author{Qirui Jiao$^{1}$, Daoyuan Chen$^{2}$, Yilun Huang$^{2}$, Bolin Ding$^{2}$, Yaliang Li$^{2}$, Ying Shen$^{1}$\\
~\\
$^{1}$Sun Yat-Sen University, $^{2}$Alibaba Group\\
\small\texttt{jiaoqr3@mail2.sysu.edu.cn, sheny76@mail.sysu.edu.cn} \\
\small\texttt{\{daoyuanchen.cdy,lielin.hyl,bolin.ding,yaliang.li\}@alibaba-inc.com}\\
}

\usepackage{multirow}
\usepackage{array}

\begin{document}

\twocolumn[{
\maketitle
\begin{center}
\centering
\includegraphics[width=\textwidth]{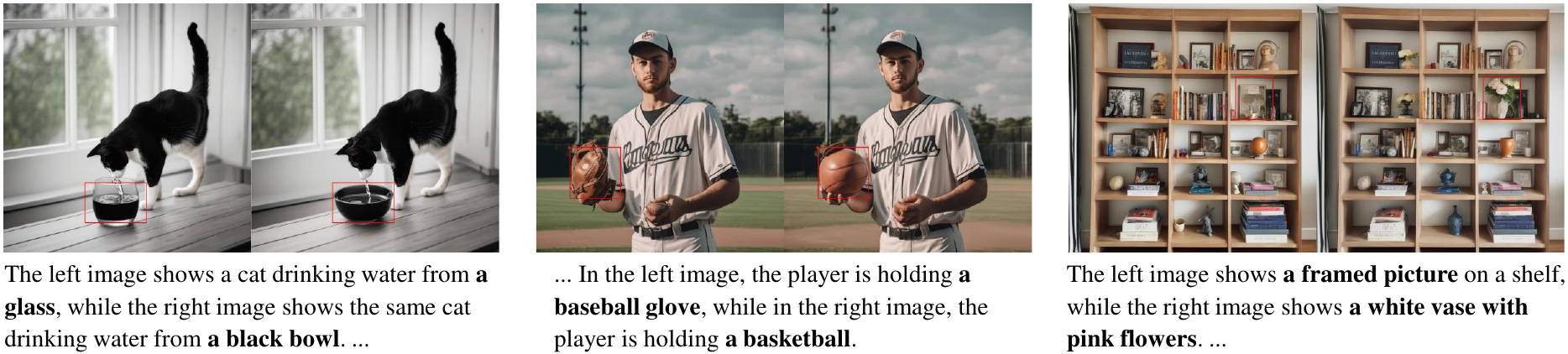}
  \captionof{figure}{Three ``object replacement'' examples within \textsc{Img-Diff}, highlighting fine-grined difference in both vision and language.}
  \label{fig:image_pair_examples_RP}
\end{center}
}]

\begin{abstract}
High-performance Multimodal Large Language Models (MLLMs) are heavily dependent on data quality. To advance fine-grained image recognition within MLLMs, we introduce a novel data synthesis method inspired by contrastive learning and image difference captioning. Our key idea involves challenging the model to discern both matching and distinct elements by scrutinizing object differences in detailed regions across similar images.
We begin by generating pairs of similar images that emphasize object variations. Following this, we employ a Difference Area Generator to pinpoint object differences, and subsequently, a Difference Captions Generator to articulate these differences. This process results in a high-quality dataset of "object replacement" samples, termed \textsc{Img-Diff}, which can be scaled as needed due to its automated nature.
We leverage this generated dataset to fine-tune state-of-the-art (SOTA) MLLMs, such as InternVL2, achieving substantial improvements across various image difference and Visual Question Answering tasks. Notably, the trained models significantly outperform existing SOTA models like GPT-4V and Gemini on the MMVP benchmark.
Additionally, we conduct comprehensive evaluations to validate the dataset's diversity, quality, and robustness, offering several insights into the synthesis of such contrastive datasets.
We release our codes and dataset to encourage further research on multimodal data synthesis and MLLMs' fundamental capabilities for image understanding.
\end{abstract}

\section{Introduction}

The emergence of large language models (LLMs) has revolutionized natural language processing \cite{gpt-3, llama}, also paved the way for the development of Multimodal Large Language Models (MLLMs) that seamlessly integrate linguistic and visual understanding. 
Improving the performance of MLLMs hinges on two primary avenues: evolving model architectures and enhancing dataset quality \cite{qin2024synergy}.
The majority of state-of-the-art (SOTA) MLLMs \cite{llava, llava1.5, llavanext, qwen-vl, mgm} employ a two-phase training strategy, commencing with a pre-training phase involving extensive image-text pairs for modality alignment, followed by a fine-tuning phase aimed at optimizing visual question answering (VQA) capabilities with specific instruction tuning datasets.

The efficacy of pre-training datasets profoundly affects MLLMs' capabilities in performing core visual tasks. 
Concurrently, the quality of visual instruction tuning datasets plays a crucial role in MLLMs' overall performance in VQA tasks and diverse downstream applications. 
With the evolution of visual instruction tuning datasets, several recent studies have explored the integration of object detection and Optical Character Recognition (OCR) datasets, such as RefCOCO \cite{refcoco}, Visual Genome \cite{vg}, OCR-VQA \cite{ocrvqa}, and TextVQA \cite{textvqa}, effectively enhancing MLLMs' proficiency in tasks requiring detailed image perception.

In this paper, we focus on a new direction for enhancing MLLM datasets, driven by the potential of object variations in image pairs to refine models' image recognition capabilities, as demonstrated by advancements in contrastive learning and image difference captioning \cite{zhang2023generalization,vixen,duda_clevr,spot}.
Specifically, we introduce a general-purpose yet challenging dataset named \textsc{Img-Diff}, which sets itself apart from existing visual instruction tuning datasets by generating pairs of highly similar images featuring subtle object alterations. 
Rather than compelling MLLMs to focus solely on a single image, our dataset challenges them to analyze paired images and articulate the differences within designated regions, meanwhile taking the high-quality textual descriptions of the difference as learning signals.
By doing so, we aim to empower MLLMs with enhanced capabilities of fine-grained image recognition.

To evaluate the effectiveness of our data synthesis method, we integrate the generated dataset into the original visual instruction tuning datasets of LLaVA-1.5 \cite{llava1.5}, MGM \cite{mgm}, and InternVL2 \cite{internvl2}, and conduct fine-tuning. 
Subsequently, we evaluate their performance on image difference benchmarks, including MMVP \cite{mmvp}, Spot-the-Diff \cite{spot}, and Image-Edit-Request \cite{image-edit-request}, as well as widely recognized MLLM benchmarks. Our evaluation results reveal that after fine-tuning with the \textsc{Img-Diff} dataset, the MLLMs achieve notable enhancements in image difference benchmarks, aligning their performance with state-of-the-art (SOTA) models. For instance, they notably surpass the SOTA models GPT-4V \cite{gpt4v} and Gemini \cite{gemini} on the MMVP benchmark, achieving an improvement of up to 12. Moreover, the models exhibit comprehensive improvements across eight well-recognized MLLM benchmarks, achieving an average score improvement of up to 3.06\%, underscoring the useful role our dataset plays in bolstering MLLMs' competencies in both image difference recognition and fine-grained image analysis.

We further evaluate the diversity and quality of our dataset, ensuring it encompasses a broad array of object categories while showcasing rich variability. 
Through meticulous manual labeling, we affirm the high quality of our dataset. 
Furthermore, we conduct ablation studies to examine the effects of various filtering intensities. 
We also investigate an alternative methodology for constructing image difference data focusing on ``object removal'', assessing its effectiveness and presenting fruitful insights on the construction of contrastive data.

Our contributions are summarized as follows:

\begin{itemize}[leftmargin=*]

\item We present a novel data synthesis method and an effect-proven \textsc{Img-Diff} dataset, comprising pairs of highly similar images, with a focus on processes such as segmentation, filtering, and detailed captioning of image differences.

\item We conduct visual instruction tuning on LLaVA-1.5-7B, MGM-7B, and InternVL2-8B using our dataset, and rigorously assess the fine-tuned models' performance on many widely-used MLLM benchmarks and image difference benchmarks. Our dataset brings substantial performance improvements to the fine-tuned MLLMs.

\item We provide a comprehensive evaluation of the diversity and quality of the generated dataset, confirming its richness and high standard. Through ablation studies, we identify good empirical filtering thresholds for such contrastive dataset.

\item We open-source our dataset and codes at \textit{\href{https://github.com/modelscope/data-juicer/tree/ImgDiff}{https://github.com/modelscope/data-juicer/tree/ImgDiff}} to facilitate ongoing research, encouraging innovative endeavors in MLLM datasets and image difference methods.

\end{itemize}

\section{Background and Related Works}

\subsection{Multimodal Large Language Models}

Multimodal Large Language Models (MLLMs) have exhibited remarkable advancements since their introduction \cite{gpt4v,llavanext,liu-etal-2024-chartthinker,gemini,mgm}. Research highlights two key factors that primarily influence the effectiveness of MLLMs: model architecture and dataset quality \cite{qin2024synergy,kdd_data_tutorial}.

With respect to model architecture, notable approaches \cite{Flamingo, idefics, idefics2, blip2, qwen-vl} leverage learnable queries to extract essential information from CLIP \cite{clip, vit} image features. Alternatively, LLaVA \cite{llava, llava1.5, llavanext} and MGM \cite{mgm} utilize projection-based interfaces to facilitate interactions between text and image modalities. Furthermore, LLaMA-Adapter \cite{LLaMA-Adapter} and LaVIN \cite{lavin} implement parameter-efficient tuning mechanisms to transfer image-related information to the LLM. A recent work also show usefullness of object detection model for MLLMs \cite{jiao2024enhancing}.

From the perspective of datasets, there are two prevalent strategies: enhancing the quality of pre-training data and improving visual instruction tuning data. The former aims for better semantic alignment between images and text by introducing abundant image-text pairs, enabling MLLMs to proficiently address fundamental visual tasks, such as image captioning. As for the latter, recent research has increasingly concentrated on refining visual instruction tuning datasets, enhancing MLLMs' performance across various question-answering tasks. Works like LLaVA \cite{llava,llava1.5,llavanext}, SPHINX \cite{SPHINX}, MGM \cite{mgm}, and InternVL2 \cite{internvl2} leverage high-quality fine-tuning datasets characterized by extensive task diversity, allowing models to excel in tasks related to image perception, reasoning, and optical character recognition. Additionally, methods such as Shikra \cite{Shikra}, ASM \cite{asm}, and PINK \cite{pink} utilize substantial amounts of object detection data to enhance the models' localization capabilities.

In contrast to previous works, our research introduces a fully automated data synthesis method and generate a dataset that emphasizes image differences, showing empirical effectiveness and great potential to augment MLLMs' VQA proficiency, object localization capabilities, and discernment of image distinctions.

\subsection{Image Difference Datasets}

Datasets focused on image differences typically consist of pairs of similar images supplemented with textual descriptions of their variations. For instance, the Spot-the-Diff dataset \cite{spot} contains pairs of street scenes captured at different times by the same surveillance cameras. The CLEVR-Change dataset \cite{duda_clevr} delineates scene variations of geometric objects against a clean backdrop. The CUB-Bird dataset \cite{birds-to-words} focuses on the nuanced differences among various bird species found in natural habitats. The Image-Edit-Request \cite{image-edit-request} dataset features edited images, accompanied by descriptions of modifications made.

Leveraging advancements in image editing technologies, some studies have employed generative models and editing techniques to create datasets centered on image differences. A prime example is InstructPix2Pix \cite{instructpix2pix}, which utilizes the Prompt-to-Prompt \cite{prompt2propmt} image editing technique to direct Stable-Diffusion-1.5 \cite{sd-1.5} in generating pairs of similar images, while employing GPT-3 \cite{gpt-3} to craft the edited text as reference captions for image differences.

Our approach, referring to InstructPix2Pix, employs the Prompt-to-Prompt technique alongside an advanced generative model Stable-Diffusion-XL \cite{sdxl}, which produces more realistic images, to generate pairs of similar images. Unlike InstructPix2Pix, we incorporate multiple filtering stages to ensure data quality, with a particular emphasis on producing difference captions that focus on specific regions rather than the entire image, which ensures greater accuracy.

\subsection{Models for Image Difference Captioning}

Image Difference Captioning (IDC) represents a specialized domain within image captioning characterized by its focus on subtle variations between images. As for the pioneering work in IDC, Spot-the-Diff \cite{spot} presents potential change clusters and employs an LSTM \cite{lstm} decoder to generate difference captions. DUDA \cite{duda_clevr} explores image differences at the semantic level, using a ResNet \cite{resnet} and an LSTM to compute dynamic attention weights and produce captions. VARD \cite{vard} introduces a viewpoint-adaptive representation disentanglement network based on LSTM for differentiating between real and pseudo changes. Meanwhile, NCT \cite{nct} employs a transformer \cite{transformers} to integrate neighboring features, and CLIP4IDC \cite{clip4idc} uses BERT-like training methodologies, adapting a CLIP model for IDC tasks. 
With the emergence of MLLMs, VIXEN \cite{vixen} marks the inaugural use of these models for IDC tasks, mapping the features of differential images to text space and employing an LLM to generate image difference captions.

Our data synthesis method and dataset is specifically designed for MLLMs. We build our data in accordance with the instruction-following format established by mainstream MLLMs such as LLaVA-1.5 and MGM, highlighting a new direction for exploration aimed at enhancing MLLMs from a data-centric perspective.

\begin{figure}[t!]
\centering
\centerline{\includegraphics[width=\columnwidth]{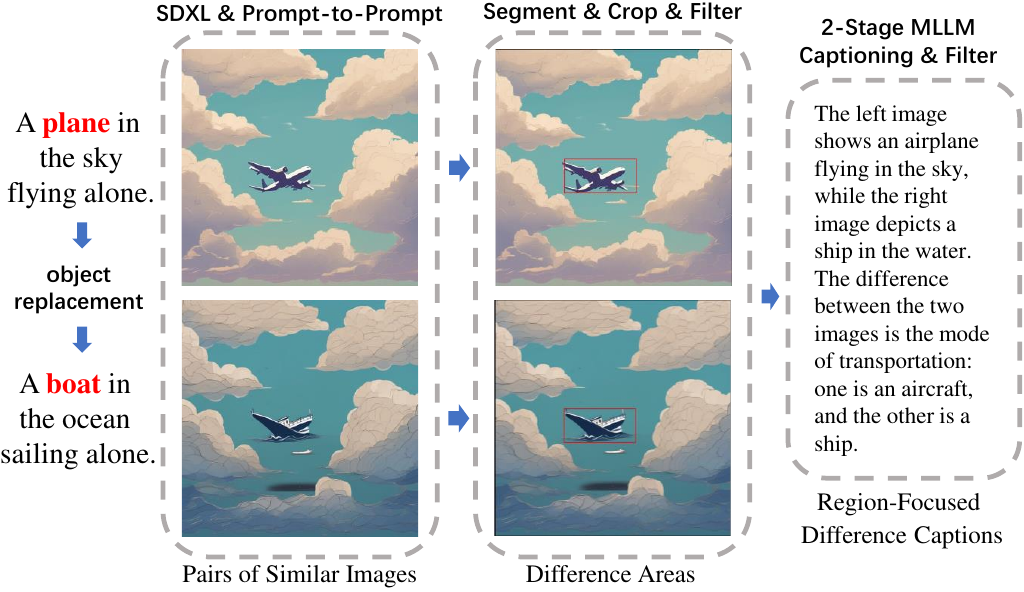}}
\caption{The generation process for ``object replacement'' data.}
\label{fig:object_replacement_overview}
\end{figure}

\begin{figure*}[t!]
\centering
\centerline{\includegraphics[width=0.76\textwidth]{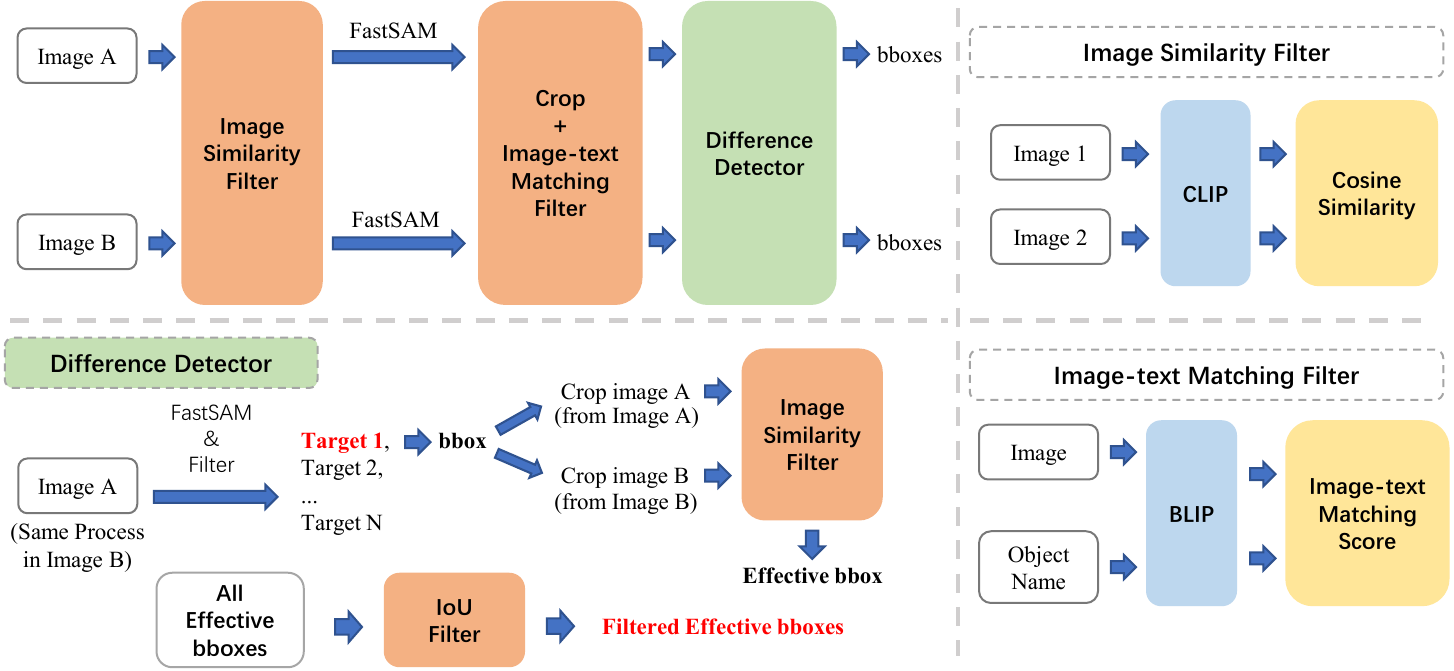}}
\caption{An overview of the Difference Area Generator and its three internal components: Image Similarity Filter, Image-text Matching Filter, and Difference Detector.}
\label{fig:difference_area_generator}
\end{figure*}

\section{Methodology}

\subsection{Overview}

In recent years, contrastive learning have significantly enhanced the image-text understanding of vision-language models \cite{clip,zhang2023generalization}. These methods typically involve constructing batches of images and texts, requiring the model to distinguish between matching and non-matching image-text pairs, which improves its ability to differentiate between semantically similar and dissimilar pairs.

Our method leverages the principles of contrastive learning to generate MLLM image-text data. Specifically, it focuses on replacing objects within image pairs, encouraging MLLMs to identify similarities and differences in specific regions. This method aims to enhance models' capacity to recognize fine-grained differences in images, guided by textual descriptions that highlight detailed distinctions.

As shown in \Cref{fig:object_replacement_overview}, the process of generating ``object replacement'' data can be divided into \textbf{three key parts}. 
The first part involves creating similar images and forming image pairs, where the only difference between the images in pairs is the objects replaced (\Cref{sec:image_pairs_generation}). The second part, named the ``\textit{Difference Area Generator}'', extracts bounding box regions that contain object differences between the images in pairs (\Cref{sec:diff_generator}). The third part, termed as the ``\textit{Difference Captions Generator}'' (\Cref{sec:caption_generator}), utilizes an MLLM to generate descriptive text for the areas with object differences and creates question-answer pairs.

This process incorporates multiple filtering operations, with specific thresholds outlined in \Cref{appendix:filtering_thres}, which we determine through experimental comparisons in \Cref{sec:ablation_study}. Additional experimental details, including model selection and the time consumption, are also presented in \Cref{appendix:additional_details}. 
For readability, we present data examples in \Cref{appendix_example}.

To enhance reproducibility and reusability, our proposed components and end-to-end construction workflow are implemented as data processing operators and configurable data recipes within Data-Juicer \cite{dj,djsandbox}.

\subsection{Image Pairs Generation}
\label{sec:image_pairs_generation}
The first step of our data synthesis method is to generate pairs of similar images as candidates. The process is shown in \Cref{fig:object_replacement_overview}. 
We employ a generative model called Stable-Diffusion-XL \cite{sdxl} and an image editing technique called Prompt-to-Prompt \cite{prompt2propmt} to generate image pairs that highlight object replacement.

We start by obtaining image captions from caption databases, which contain descriptions biased towards real photos. Then, we use an LLM to perform object replacement in the captions. The prompt used is ``\textit{Here is a sentence: `INPUT'. Please only replace one of the objects in this sentence with another object.}'' Here, \textit{INPUT} refers to the original caption. Next, based on the caption pairs, we use the Stable-Diffusion-XL and the Prompt-to-Prompt to generate image pairs with only few objects replaced.

\begin{figure*}[t!]
\centering
\centerline{\includegraphics[width=0.9\textwidth]{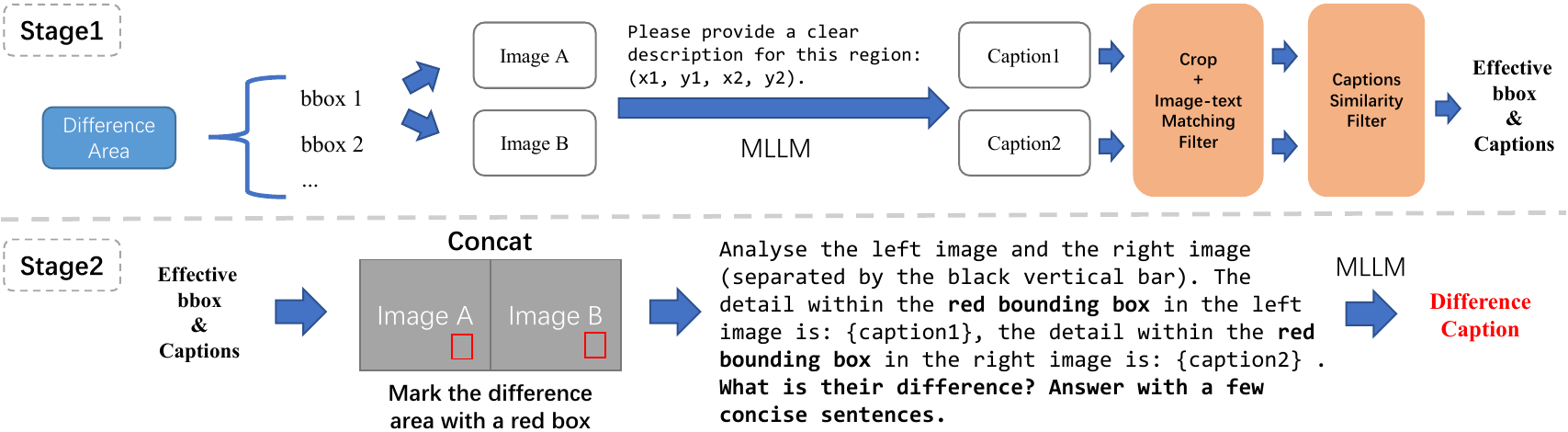}}
\caption{An overview of the Difference Captions Generator and its two stages.}
\label{fig:difference_captions_generator}
\end{figure*}

\subsection{Difference Area Generator}
\label{sec:diff_generator}
\subsubsection{Overview}
The Difference Area Generator aims to identify the locations of object differences between the two images in pairs. Although object detection models are capable of detecting objects in images, the range of object categories is quite limited \cite{DINO, yolov8}. Therefore, to increase the number of detectable object categories and enhance dataset diversity, we develop the Difference Area Generator based on segmentation and image similarity comparisons.
The process is illustrated in \Cref{fig:difference_area_generator}. 

We first use an Image Similarity Filter to obtain image pairs with high similarity but not identical. Next, we use the FastSAM \cite{fastsam} to perform image segmentation on each image. After that, we crop the images based on the bounding box information obtained from segmentation and use an Image-text Matching Filter to filter the cropped sub-images for the presence of valid objects. Finally, we use a Difference Detector to determine whether there are indeed differences between the bounding box regions of the two images and perform IoU (Intersection over Union) filtering to remove overlapping bounding boxes, ultimately obtaining valid bounding box information.

\subsubsection{Image Similarity Filter} 
The Image Similarity Filter aims to filter image pairs based on the degree of similarity. The module first uses CLIP \cite{clip} to extract image features from each image in pairs and then calculates the cosine similarity score. If their score falls within the pre-set threshold, the image pair will be considered valid. Specifically, we use the module twice in the Difference Area Generator. Before the segmentation, we use the module to ensure that the images in pairs are highly similar but not the same. In the Difference Detector stage, after cropping, we use the module to filter the sub-image pairs and keep only the differing ones.

\subsubsection{Image-text Matching Filter}

The Image-text Matching Filter determines whether an image contains valid objects (i.e. the replaced or replacing objects). This module first uses BLIP \cite{blip} to extract image features, which are then compared with textual features of object names. When the image-text matching score falls within the pre-set threshold, we consider the image to contain valid objects. 
In the mid-stage of the Difference Area Generator, after cropping, we use the module to determine whether these sub-images contain valid objects and get valid bounding boxes.

\subsubsection{Difference Detector}

The Difference Detector determines whether there are differences between the bounding box regions of the two images. Based on a given bounding box, we first crop two sub-images from both image A and image B. The two sub-images are then filtered through the Image Similarity Filter and the bounding box is considered effective only if the difference is significant enough. After processing all bounding boxes, we use the IoU method to filter out overlapping bounding boxes. Only the bounding boxes with a higher degree of difference are retained.

\subsection{Difference Captions Generator}
\label{sec:caption_generator}
\subsubsection{Overview}
After obtaining the valid bounding box regions, we use the Difference Captions Generator to generate descriptions for the differences inside these areas (with each round of the process focusing on only one bounding box in one image pair). Evidently, an image pair can contain multiple object differences but a single difference caption cannot fully capture all of them. Therefore, we highlight specific regions with red boxes and provide targeted difference captions to ensure greater accuracy.

The module consists of two stages: the first stage generates captions for the content in the bounding box regions and then filters the bounding boxes using an Image-text Matching Filter and a Captions Similarity Filter. The second stage uses the content captions and the images highlighted with red boxes to generate difference captions. The overview is shown in \Cref{fig:difference_captions_generator}.

\subsubsection{Stage1: Object Labeling \& Filtering}

In Stage 1, for each image pair, we first select N bounding box regions with the lowest similarity between images (N is set to 5 in this project) as candidate regions. Then, for each bounding box, we use the MLLM LLaVA-NEXT \cite{llavanext} to describe its corresponding regions and then apply two filtering processes: the first filter is an Image-text Matching Filter, which checks whether the content of the regions matches the captions; the second filter is an Captions Similarity Filter, which assesses whether there are differences between the two captions. Once the filtering is complete, we obtain valid bounding boxes and captions for subsequent difference captioning.

\subsubsection{Captions Similarity Filter}

The Captions Similarity Filter determines whether the two captions of the same bounding box coordinates are different. We use CLIP to obtain text features and calculate the cosine similarity between them. When the score is low enough, we consider the two captions to be different.

\subsubsection{Stage2: Difference Captions Generating}

In Stage 2, for each valid bounding box of each image pair, we first draw two red boxes into the images based on the bounding box coordinates, highlighting the differences for easier localization. Then, we provide the MLLM LLaVA-NEXT with the captions of the bounding box content and instruct it to describe the difference based on these captions and the highlighted images. Finally, we can obtain the difference caption for the highlighted area.

\subsection{Data Statistics}
Using captions from MSCOCO \cite{mscoco}, we generate 118K pairs of similar images. We then employ the Image Similarity Filter to get 38,533 highly similar but not identical image pairs. Next, we use the Difference Area Generator to filter and produce 117,779 pieces of valid bounding box information (with a maximum of 5 valid bounding boxes per image pair). Finally, we employ the Difference Captions Generator to filter and generate 12,688 high-quality ``object replacement'' instances. Our evaluations on the main page are based on this dataset.

In addition, we also use captions from the LLaVA pre-training dataset to generate 34,538 ``object replacement'' samples. In \Cref{sec:appendix_data_size} of the appendix, we evaluate this dataset and discuss the relationship between data quantity, data quality, and model performance improvement, emphasizing the importance of prioritizing quality over quantity.
\section{Evaluation and Main Results}

\subsection{Evaluation Settings}
To evaluate the effectiveness of the \textsc{Img-Diff} dataset, we use it to fine-tune SOTA MLLMs, including LLaVA-1.5-7B, MGM-7B, and InternVL2-8B on the main page, as well as InternVL2-1B and LLaVA-1.5-13B in the supplementary materials. We evaluate these models on extensive benchmarks commonly used for image difference and MLLMs. Specifically, the image difference benchmarks include MMVP \cite{mmvp}, Spot-the-Diff \cite{spot}, and Image-Edit-Request \cite{image-edit-request}. Besides, the details and results of the MLLM benchmarks are shown in \Cref{sec:mllm_benchmarks}.

During our fine-tuning process, we first mix our data with MLLMs' original visual instruction tuning data respectively. For LLaVA-1.5 and MGM, we conduct the fine-tuning anew. For InternVL2, we follow the guidelines from its official repository and perform a second fine-tuning. (To ensure a fair comparison, we perform a second fine-tuning on InternVL2 using its original fine-tuning dataset, termed as InternVL2-FT, and use it as a baseline.) Regarding the Spot-the-Diff and Image-Edit-Request benchmarks, as they contain data of training splits, we further fine-tune the fine-tuned MLLMs for an additional 2 epochs using only these benchmarks' training data.

In the tables, ``RP'' represents ``object replacement'' data.

\subsection{Results on the MMVP Benchmark}
The MMVP benchmark is designed to systematically assess the visual capabilities of MLLMs. Its data processing method is highly related to image difference: it first collects CLIP-blind pairs, which have similar CLIP features but differ in image content. Then, the differences between the images are manually described and question-answer pairs are created. Hence, the questions in MMVP is highly relevant to our dataset, as both place significant emphasis on the differences between similar images.

\begin{figure}[htbp]
\centering
\centerline{\includegraphics[width=\columnwidth]{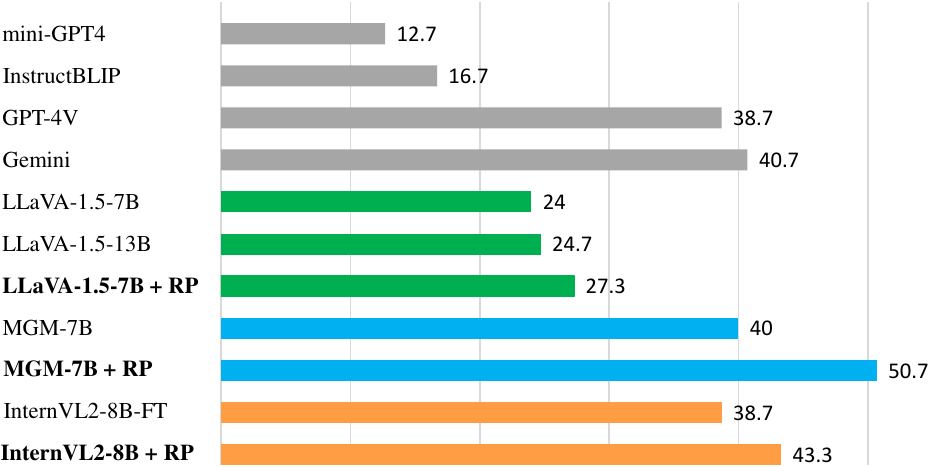}}
\caption{Performance comparison on the MMVP benchmark.}
\label{tab:mmvp}
\end{figure}

As shown in \Cref{tab:mmvp}, fine-tuning MLLMs with our ``object replacement'' data significantly improves their performance on the MMVP benchmark. After fine-tuning, the score of LLaVA-1.5-7B exceeds that of LLaVA-1.5-13B. Furthermore, the fine-tuned MGM-7B shows a significant improvement in score compared to the original MGM-7B, even surpassing the scores of the SOTA models GPT-4V and Gemini by up to 12. Additionally, the performance of InternVL2-8B is also enhanced.
These results suggest that our dataset enhances MLLMs' abilities to distinguish images with similar CLIP features but different content.

\subsection{Results on the Spot-the-Diff Benchmark}
The dataset of Spot-the-Diff comprises pairs of street view images that display subtle object differences. These images are obtained by capturing scenes from fixed surveillance cameras at different time.

% % Please add the following required packages to your document preamble:
% % \usepackage{booktabs}
% % \usepackage{graphicx}
% % \usepackage[normalem]{ulem}
% % \useunder{\uline}{\ul}{}
% \begin{table}[h]
% \centering
% \caption{Performance comparison on Spot-the-Diff.}
% \label{tab:spot-the-diff}
% \resizebox{0.8\columnwidth}{!}{%
% \begin{tabular}{@{}ccccc@{}}
% \toprule
% Model                    & BLEU          & METEOR        & CIDEr-D       & ROUGE-L     \\ \midrule
% VAM\cite{vam}                      & \underline{10.1}          & 12.4          & 38.1          & 31.3        \\
% IFDC\cite{ifdc}                     & 8.7           & 11.7          & 37            & 30.2        \\
% DUDA+Aux\cite{duda_aux}                 & 8.1           & 12.5          & 34.5          & 29.9        \\
% VACC\cite{vacc}                     & 9.7           & 12.6          & 41.5          & {\underline{32.1}}        \\
% \midrule
% LLaVA-1.5-7B             & 8.5           & 12.0          & 38.3          & 30.1        \\
% \textbf{LLaVA-1.5-7B + RP} & 9.7           & {\underline{13.0}}          & 43.2          & 30.8        \\ \midrule
% MGM-7B                      & 9.9           & 12            & {\underline{46.3}}          & 31.5        \\
% \textbf{MGM-7B + RP}          & \textbf{10.8}    & \textbf{13.1}    & \textbf{53.5} & \textbf{33.0}  \\ \bottomrule
% \end{tabular}%
% }
% \end{table}

% \usepackage{booktabs}
% \usepackage{graphicx}
\begin{table}[h]
\centering
\caption{Performance comparison on Spot-the-Diff.}
\label{tab:spot-the-diff}
\resizebox{0.88\columnwidth}{!}{%
\begin{tabular}{@{}ccccc@{}}
\toprule
Model                      & BLEU             & METEOR           & CIDEr-D          & ROUGE-L          \\ \midrule
VAM\cite{vam}              & \underline{10.1} & 12.4             & 38.1             & 31.3             \\
IFDC\cite{ifdc}            & 8.7              & 11.7             & 37               & 30.2             \\
DUDA+Aux\cite{duda_aux}    & 8.1              & 12.5             & 34.5             & 29.9             \\
VACC\cite{vacc}            & 9.7              & 12.6             & 41.5             & \underline{32.1} \\ \midrule
LLaVA-1.5-7B               & 8.5              & 12.0             & 38.3             & 30.1             \\
\textbf{LLaVA-1.5-7B + RP} & 9.7              & \underline{13.0} & 43.2             & 30.8             \\ \midrule
MGM-7B                     & 9.9              & 12.0             & \underline{46.3} & 31.5             \\
\textbf{MGM-7B + RP}       & \textbf{10.8}    & \textbf{13.1}    & \textbf{53.5}    & \textbf{33.0}    \\ \midrule
InternVL2-8B-FT        & 6.6              & 11.7             & 26.5             & 27.3             \\
\textbf{InternVL2-8B + RP} & 8.4              & 12.8             & 32.2             & 28.5             \\ \bottomrule
\end{tabular}%
}
\end{table}

Following previous works, our fine-tuned MLLMs are evaluated on BLEU \cite{bleu}, METEOR \cite{meteor}, CIDEr-D \cite{cider} and ROUGE-L \cite{rouge}. As shown in \Cref{tab:spot-the-diff}, after fine-tuning with our ``object replacement'' data, both LLaVA-1.5-7B, MGM-7B, and InternVL2-8B achieve significant performance gains on Spot-the-Diff. Compared to score increases seen in prior models, our dataset yields even greater enhancements than those from previous iterations of image difference models. These positive results indicate that our dataset can enhance the ability of MLLMs to detect subtle differences between similar images.

\subsection{Results on Image-Editing-Request}
The Image-Editing-Request benchmark focuses on image editing. Each instance in its dataset consists of an image pair (i.e. a source image and an edited image) and an editing instruction that describes the transformation. During the evaluation, our models are required to generate transformation descriptions for these image pairs, and we then calculate the BLEU, METEOR, CIDEr-D, and ROUGE-L scores using the models' responses and the reference answers.

% % Please add the following required packages to your document preamble:
% % \usepackage{booktabs}
% % \usepackage{graphicx}
% % \usepackage[normalem]{ulem}
% % \useunder{\uline}{\ul}{}
% \begin{table}[h]
% \centering
% \caption{Performance comparison on Image-Edit-Request.}
% \label{tab:image-edit-request}
% \resizebox{0.8\columnwidth}{!}{%
% \begin{tabular}{@{}ccccc@{}}
% \toprule
% Model           & BLEU          & METEOR        & CIDEr-D       & ROUGE-L       \\ \midrule
% VARD\cite{vard}            & 10            & 14.8          & 35.7          & 39            \\
% CLIP4IDC\cite{clip4idc}        & 8.2           & 14.6          & 32.2          & 40.4          \\
% NCT\cite{nct}             & 8.1           & 15            & 34.2          & 38.8          \\
% BiDiff\cite{bidiff}          & 6.9           & 14.6          & 27.7          & 38.5          \\
% VIXEN\cite{vixen}           & 8.6           & 15.4          & 38.1          & 42.5          \\ \midrule
% LLaVA-1.5-7B    & 15.1          & 17.8          & 60.6          & 45.2          \\
% \textbf{LLaVA-1.5-7B + RP} & 16.2    & \textbf{19.5} & 60.9          & \textbf{46.7} \\ \midrule
% MGM-7B             & \underline{16.5} & 17.7          & {\underline{66.8}}    & 44.8          \\
% \textbf{MGM-7B + RP}          & \textbf{16.6} & {\underline{18.2}}    & \textbf{68.1} & {\underline{45.7}}    \\ \bottomrule
% \end{tabular}%
% }
% \end{table}

% \usepackage{booktabs}
% \usepackage{graphicx}
\begin{table}[h]
\centering
\caption{Performance comparison on Image-Edit-Request.}
\label{tab:image-edit-request}
\resizebox{0.9\columnwidth}{!}{%
\begin{tabular}{@{}ccccc@{}}
\toprule
Model                      & BLEU           & METEOR         & CIDEr-D        & ROUGE-L        \\ \midrule
VARD\cite{vard}            & 10             & 14.8           & 35.7           & 39             \\
CLIP4IDC\cite{clip4idc}    & 8.2            & 14.6           & 32.2           & 40.4           \\
NCT\cite{nct}              & 8.1            & 15             & 34.2           & 38.8           \\
BiDiff\cite{bidiff}        & 6.9            & 14.6           & 27.7           & 38.5           \\
VIXEN\cite{vixen}          & 8.6            & 15.4           & 38.1           & 42.5           \\ \midrule
LLaVA-1.5-7B               & 15.1           & 17.8           & 60.6           & 45.2           \\
\textbf{LLaVA-1.5-7B + RP} & 16.2           & \textbf{19.5}  & 60.9           & \textbf{46.7}  \\ \midrule
MGM-7B                     & \underline{16.5} & 17.7           & \underline{66.8} & 44.8           \\
\textbf{MGM-7B + RP}       & \textbf{16.6}  & \underline{18.2} & \textbf{68.1}  & \underline{45.7} \\ \midrule
InternVL2-8B-FT       & 12.4           & 14.1           & 51.5           & 38.9           \\
\textbf{InternVL2-8B + RP} & 12.5           & 14.2           & 56.0           & 39.4           \\ \bottomrule
\end{tabular}%
}
\end{table}

As shown in \Cref{tab:image-edit-request}, LLaVA-1.5-7B, MGM-7B, and InternVL2-8B originally show SOTA performance on the Image-Edit-Request benchmark. Upon incorporating our ``object replacement'' data for better fine-tuning, the performance of all three models improves even more, achieving new SOTA scores. This suggests that our dataset enhances MLLMs' abilities to recognize similarities and dissimilarities in image pairs, as well as enables them to describe differences more accurately.

\subsection{Results on MLLM Benchmarks}
\label{sec:mllm_benchmarks}

Aside from the evaluations related to image difference discrimination, we also assess the performance of our ``object replacement'' data in enhancing the comprehensive abilities of MLLMs. We test the fine-tuned MLLMs using commonly used MLLM benchmarks, including VQAv2 \cite{vqav2} and GQA \cite{gqa} for assessing the comprehensive VQA capabilities; MMBench \cite{mmbench}, MM-Vet \cite{mmvet}, ScienceQA \cite{scienceqa}, and SEED-Bench \cite{seed} for testing perceptual and reasoning abilities; and POPE \cite{pope} for evaluating fine-grained object localization abilities. \Cref{tab:mllm_benchmarks} presents the results on these MLLM benchmarks, with the $\triangle$ metric indicating the percentage improvement averaged across them.

\begin{table}[h]
\centering
\caption{Performance comparison on 8 MLLM benchmarks.}
\label{tab:mllm_benchmarks}
\resizebox{0.95\columnwidth}{!}{%
\begin{tabular}{@{}cccccc@{}}
\toprule
Model                      & VQA$^{v2}$                       & GQA                              & POPE                                 & MMB                                  & MMB$^{CN}$                           \\ \midrule
LLaVA-1.5-7B               & 78.5                             & 62                               & 85.9                                 & 64.3                                 & 58.3                                 \\
\textbf{LLaVA-1.5-7B + RP} & 79.3 \textcolor{red}{$\uparrow$} & 62.8 \textcolor{red}{$\uparrow$} & 86.4 \textcolor{red}{$\uparrow$}     & 66.1 \textcolor{red}{$\uparrow$}     & 59.8 \textcolor{red}{$\uparrow$}     \\ \midrule
MGM-7B                     & 80.4                             & 62.6                             & 86                                   & 69.3                                 & 58.9                                 \\
\textbf{MGM-7B + RP}       & 80.7 \textcolor{red}{$\uparrow$} & 62.7 \textcolor{red}{$\uparrow$} & 86.2 \textcolor{red}{$\uparrow$}     & 68.7 \textcolor{green}{$\downarrow$} & 59.6 \textcolor{red}{$\uparrow$}     \\ \midrule
InternVL2-8B-FT        & 81.8                             & 62.6                             & 87.7                                 & 82.5                                 & 81.5                                 \\
\textbf{InternVL2-8B + RP} & 81.8                             & 62.6                             & 88.0 \textcolor{red}{$\uparrow$}       & 82.7 \textcolor{red}{$\uparrow$}     & 81.4 \textcolor{green}{$\downarrow$} \\ \bottomrule
\multicolumn{1}{l}{}       & \multicolumn{1}{l}{}             & \multicolumn{1}{l}{}             & \multicolumn{1}{l}{}                 & \multicolumn{1}{l}{}                 & \multicolumn{1}{l}{}                 \\ \toprule
Model                      & MM-Vet                           & SQA$^{I}$                        & SEED                                 & $\triangle$                          &                                      \\ \midrule
LLaVA-1.5-7B               & 30.5                             & 66.8                             & 58.6                                 & -                                    &                                      \\
\textbf{LLaVA-1.5-7B + RP} & 33.2 \textcolor{red}{$\uparrow$} & 68.2 \textcolor{red}{$\uparrow$} & 61.7 \textcolor{red}{$\uparrow$}     & \textcolor{red}{$+3.06\%$}           &                                      \\ \midrule
MGM-7B                     & 40.8                             & 70.6                             & 63.5                                 & -                                    &                                      \\
\textbf{MGM-7B + RP}       & 44.1 \textcolor{red}{$\uparrow$} & 71.7 \textcolor{red}{$\uparrow$} & 63.2 \textcolor{green}{$\downarrow$} & \textcolor{red}{$+1.28\%$}           &                                      \\ \midrule
InternVL2-8B-FT        & 49.2                             & 96.5                             & 69.5                                 & -                                    & \multicolumn{1}{l}{}                 \\
\textbf{InternVL2-8B + RP} & 52.6 \textcolor{red}{$\uparrow$} & 96.6 \textcolor{red}{$\uparrow$} & 69.9 \textcolor{red}{$\uparrow$}     & \textcolor{red}{$+1.01\%$}           & \multicolumn{1}{l}{}                 \\ \bottomrule
\end{tabular}%
}
\end{table}

Based on \Cref{tab:mllm_benchmarks}, after fine-tuning with our dataset, the performance of LLaVA-1.5-7B shows a comprehensive improvement, with an average increase of 3.06\% across all benchmarks.
For MGM-7B and InternVL2-8B, the improvements brought by our dataset are not as pronounced as those observed with LLaVA-1.5-7B, as their training datasets already encompass a large volume of high-quality data, but they still achieve an average increase of 1.28\% and 1.01\%. These improvements indicate that fine-tuning MLLMs with our dataset not only enhances their ability to discern differences but also improves their overall visual capabilities, thereby making them better address VQA tasks.

\section{Evaluation of Data Quality and Diversity}
\subsection{Data Quality}
\begin{figure}[htbp]
\centering
\centerline{\includegraphics[width=\columnwidth]{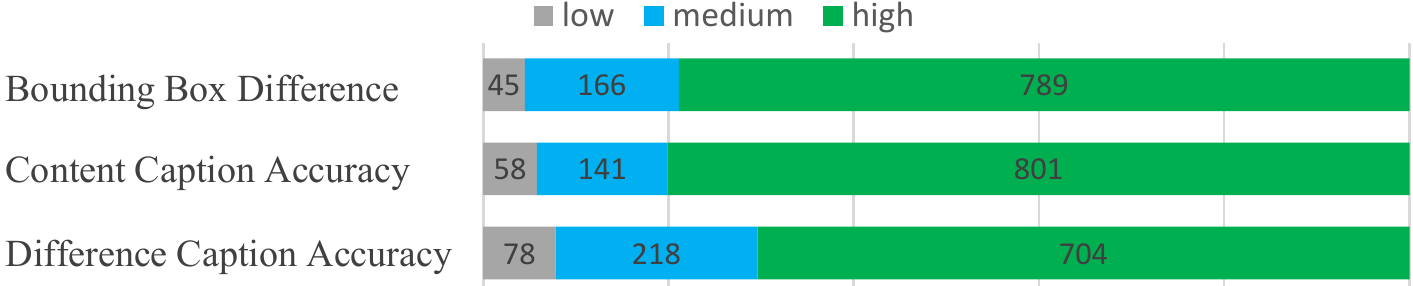}}
\caption{Quality assessment of the ``object replacement'' data.}
\label{tab:quality_table}
\end{figure}

To assess the quality of the \textsc{Img-Diff} dataset, we randomly select 1,000 instances of ``object replacement'' data and employ multiple professional dataset annotators to evaluate the samples based on three metrics. The final scores are determined through a voting process.
Specifically, the first metric is ``\textbf{Bounding Box Difference}'', which evaluates whether there are differences between the two highlighted regions in pairs. If the objects are different, we score it as ``high''; if the objects are the same but their features (such as color, shape, etc.) are different, we score it as ``medium''; if the objects are the same and their features are similar, we score it as ``low''. The second metric is ``\textbf{Content Caption Accuracy}'', which evaluates whether the captions generated by Stage 1 of the Difference Captions Generator accurately describe the two highlighted regions. If the captions are correct, we score it as ``high''; if the captions identify the objects but incorrectly describe their features, we score it as ``medium''; if the captions incorrectly identify the objects, we score it as ``low''. The third metric is ``\textbf{Difference Caption Accuracy}'', which evaluates whether the final difference captions accurately describe the object differences between highlighted regions of the image pairs.  If the description is accurate, we score it as ``high''; if it recognizes the objects but the feature description is incorrect, we score it as ``medium''; if the object recognition is incorrect, we score it as ``low''. The results are shown in \Cref{tab:quality_table}.

Based on \Cref{tab:quality_table}, our dataset demonstrates a high level of quality. For the ``Bounding Box Difference'' metric, only 4.5\% of the instances are classified as ``low'', and nearly 80\% of instances exhibit valid object differences between their two highlighted regions. In terms of ``Content Caption Accuracy'', 80.1\% of highlighted region pairs are described accurately, indicating that using an MLLM for labeling is effective and that our filtering strategy is also functioning well. For the ``Difference Caption Accuracy'' metric, over 70\% of the difference descriptions are completely accurate, with 21.8\% of the samples having errors solely in feature labeling while still maintaining correct descriptions of object differences, which underscores the effectiveness of our difference caption generation strategy.

\subsection{Data Diversity}

In designing the generation process, we have made efforts to enhance the diversity of our dataset, which can be divided into two aspect: (a) the inherent diversity of the caption databases; and (b) the introduction of new object names through our object replacement strategy. As for the former, we are using the captions from MS COCO and the LLaVA pre-training dataset, which cover a majority of object categories. We can easily further enhance the data diversity by employing caption databases with greater variability. As for the latter, we employ two methods, including increasing the temperature of the LLM used for noun replacement, as well as randomly replacing object names with nouns from a noun lexicon (shown in Appendix, \Cref{sec:appendix_diversity_lexicon}). We can further improve data diversity by expanding the noun lexicon.

By analyzing the valid object names included in the captions that describe highlighted regions, we assess the diversity of our ``object replacement'' data. Specifically, we count the total number of object categories covered, and the total number of unique ``object replacement pairs''.
Through our statistical analysis, we find that our dataset covers 1,203 object categories, which encompasses most real-life objects. An ``object replacement pair'' refers to the combination of the replaced and replacing object names. Our dataset includes 3,680 unique ``object replacement pairs''.

To evaluate the coverage of common object categories in our dataset, we analyze the occurrences of the object names from the Object365 \cite{object365} dataset in our own dataset. The results show that the object names from the Object365 dataset appears 13,164 times in total, which accounts for approximately 52.06\% of the total occurrences of object names. 

These statistics show that our data covers a substantial number of common object names, ensuring a high frequency of common objects. Additionally, less common object names make up nearly half of our dataset, indicating that our dataset is both diverse and comprehensive.

\section{More Empirical Supports and Details}

We list additional experimental details and extensions of our data synthesis method in the supplementary material, and conduct further experiments to validate the effectiveness of the \textsc{Img-Diff} dataset.

\paragraph{Further Experiments and Analysis.} In \Cref{sec:appendix_other_image_difference_dataset}, we compare the \textsc{Img-Diff} dataset with existing image difference datasets on characteristics and performance, validating its superiority. In \Cref{sec:appendix_data_size}, we discuss the relationship between the data volume and model performance improvement. By appropriately expanding the dataset, we achieve further improvement in model performance. In \Cref{sec:appendix_diversity_lexicon}, 
we explore the use of lexicons for object replacement and validate its effectiveness in enhancing diversity. In \Cref{appendix_cocot}, we evaluate our dataset using the Contrastive Chain-of-Thought method, further confirming that our dataset enhances the model's ability to recognize image differences. In \Cref{sec:appendix_other_mllms}, we validate the effectiveness of the \textsc{Img-Diff} dataset on MLLMs of different scales.

\paragraph{Implementation Details.} In \Cref{sec:ablation_study}, we investigate the effect of different filtering intensities on the performance of the generated datasets. In \Cref{appendix:additional_details}, we present additional experimental details, including image pair preprocessing, model training procedures, model selection, filtering thresholds, and time consumption.

\paragraph{The ``Object Removal'' Exploration.} In \Cref{appendix_rm_data}, we generate an extended dataset that focuses on object removal, which prompts MLLMs to analyze which image contains a specific object. The new data brings further improvement to the fine-tuned MLLM. 
\section{Conclusion}

In this paper, we draw inspiration from recent advances in contrastive learning and image difference captioning to propose a novel method of contrastive data synthesis. This method enables the creation of a high-quality dataset called \textsc{Img-Diff}, which highlights object differences focusing on fine-grained regions in images. Specifically, we first generate pairs of similar images where the main focus is on object replacement. Then, we use the proposed \textit{Difference Area Generator} and \textit{Difference Captions Generator} to generate difference captions and form question-answer pairs. In contrast to previous image difference datasets, our dataset focuses exclusively on specific regions inside images. This characteristic circumvents the issue where a single description cannot fully capture the differences of the whole images in pairs, enhancing the accuracy. Afterwards, we fine-tune many MLLMs using the generated dataset, yielding high-performance scores on par with SOTA models in image difference tasks and comprehensive performance improvements in eight widely recognized MLLM benchmarks. These results confirm the effectiveness of our dataset to improve the ability of MLLMs in recognizing detailed differences between images.

In a nutshell, we provide a series of insights about the construction of high-quality image difference datasets, showing great potential to effectively and efficiently enhance MLLMs via contrastive data-centric approaches. 
With this work, we hope it can catalyze further investigation into the realm of image difference datasets and the fine-grained image recognition capabilities of MLLMs.
{
    \small
    \bibliographystyle{ieeenat_fullname}
    \bibliography{main}
}

\clearpage
\setcounter{page}{1}
\maketitlesupplementary

\section{Overview}
We provide more details and experiments of this work in the supplementary material and organize them as follows:
\begin{itemize}

    \item \Cref{sec:appendix_other_image_difference_dataset}. \textbf{Comparison with Existing Image Difference Datasets}: We compare the \textsc{Img-Diff} dataset with existing image difference datasets in terms of characteristics and performance, highlighting the advantages of our dataset.

    \item \Cref{sec:appendix_data_size}. \textbf{Prioritizing Quality Over Quantity}: We clarify that our choice to use 13K samples for testing is motivated by the typical size of task-specific datasets used for MLLM fine-tuning. Furthermore, by expanding the dataset to four times its original size, we confirm that the relationship between data size and performance gains is not linear.

    \item \Cref{sec:appendix_diversity_lexicon}. \textbf{Expanding Diversity with Lexicons}: We use a lexicon to generate object replacement data and test the new dataset. The results validate the effectiveness of this lexicon-based strategy in enhancing data diversity.

    \item \Cref{appendix_cocot}. \textbf{Performance Based on Contrastive Chain-of-Thought}: We evaluate our dataset using the Contrastive Chain-of-Thought method. The results confirm that our dataset enables the fine-tuned model to more accurately describe image differences, thereby enhancing the model's VQA capability.

    \item \Cref{sec:appendix_other_mllms}. \textbf{Testing on MLLMs at Different Scales}: We test the performance of the \textsc{Img-Diff} dataset across MLLMs of different scales. The results indicate that the performance gains brought by our dataset are not limited by scale.

    \item \Cref{sec:ablation_study}. \textbf{Ablation Studies}: We explore the impact of varying filter intensities on the performance of the final dataset. As a result, we identify an optimal threshold that balances data quality and quantity.

    \item \Cref{appendix:additional_details}. \textbf{Additional Details of Experiments}: We present additional details, including the preprocessing methods for image pairs, the standard training strategies for MLLMs, the model selection and rationale behind our approach, the filtering thresholds applied throughout the work, and the time consumption for generating data.

    \item \Cref{appendix_rm_data}.  \textbf{The ``Object Removal'' Exploration}: We generate an extended dataset that focuses on object removal. Additionally, we experimentally validate its effectiveness.

    \item \Cref{appendix_example}. \textbf{Examples}: We present several examples of our ``object replacement'' data and ``object removal'' data, highlighting detailed information.

\end{itemize}

\section{Comparison with Existing Image Difference Datasets}
\label{sec:appendix_other_image_difference_dataset}
\subsection{Characteristics Comparison}
\begin{table*}[t]
\caption{Comparison of different image difference datasets. ``Open-Domain'' refers to whether the dataset has a limited or unrestricted range of object coverage; ``Automatic'' indicates whether the dataset can be fully generated through automation without human intervention; and ``Region-Focused'' describes whether the dataset emphasizes detailed regions rather than the overall image.
}
\centering
\resizebox{\textwidth}{!}{

\begin{tabular}{m{0.1\linewidth} m{0.1\linewidth} m{0.1\linewidth} m{0.1\linewidth} m{0.1\linewidth} m{0.15\linewidth} m{0.15\linewidth} m{0.25\linewidth}}
\toprule
\multicolumn{1}{c}{\textbf{Datasets}} & \multicolumn{1}{c}{\textbf{Open-Domain?}} & \multicolumn{1}{c}{\textbf{Automatic?}} & \multicolumn{1}{c}{\textbf{Region-Focused?}} & \multicolumn{1}{c}{\textbf{Size}}                                     & \multicolumn{1}{c}{\textbf{Source}} & \multicolumn{1}{c}{\textbf{Target}}                                           & \multicolumn{1}{c}{\textbf{Text}} \\ \midrule

\multicolumn{1}{c}{CUB-Bird \cite{birds-to-words}}             & \multicolumn{1}{c}{$\times$}   & \multicolumn{1}{c}{$\times$} & \multicolumn{1}{c}{$\times$}         & \multicolumn{1}{c}{11,788}                             & \includegraphics[width=0.15\textwidth]{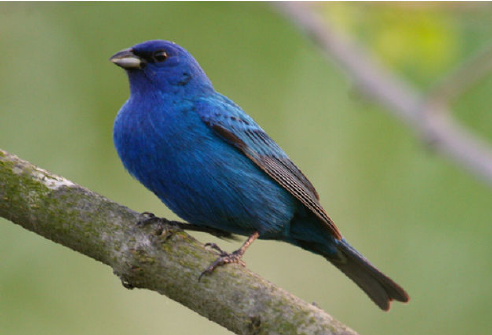}      & \includegraphics[width=0.15\textwidth]{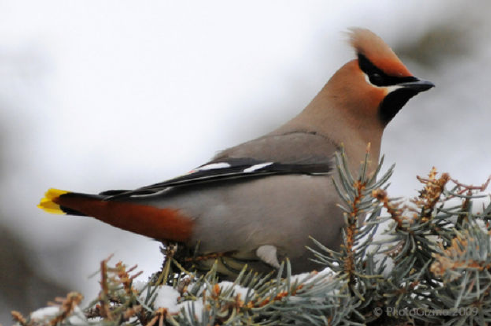}           & \textit{``This is a grey bird with a brown and yellow tail wing and a red head. (Select)''}  \\

\multicolumn{1}{c}{Spot-the-Diff \cite{spot}}                      & \multicolumn{1}{c}{$\times$}   & \multicolumn{1}{c}{$\times$}  & \multicolumn{1}{c}{$\times$}  & \multicolumn{1}{c}{13,192}                               & \includegraphics[width=0.15\textwidth]{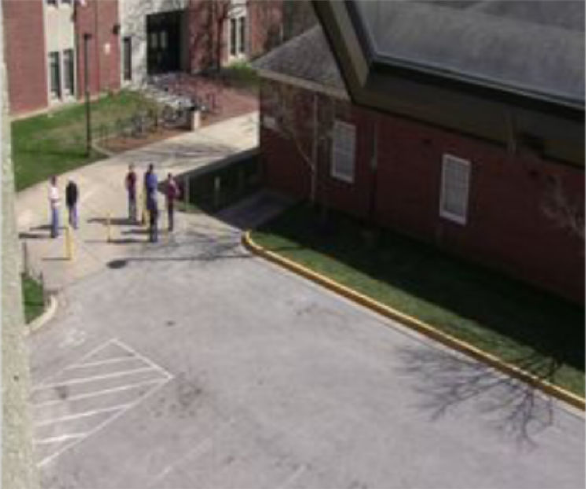}      & \includegraphics[width=0.15\textwidth]{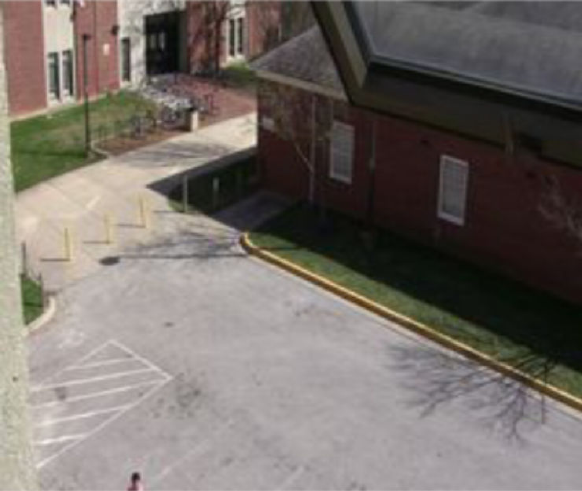}                           &  \textit{``The people in the parking lot are no longer there.''}     \\

\multicolumn{1}{c}{Image-Edit-Request \cite{image-edit-request}}                                              & \multicolumn{1}{c}{\checkmark}   & \multicolumn{1}{c}{$\times$} & \multicolumn{1}{c}{$\times$}               & \multicolumn{1}{c}{3,939}                                & \includegraphics[width=0.15\textwidth]{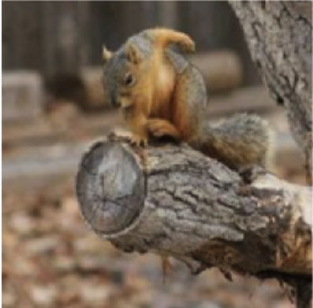}      & \includegraphics[width=0.15\textwidth]{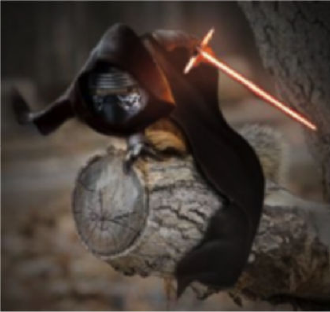} & \textit{``Add a sword and a cloak to the squirrel.''}      \\ 
\multicolumn{1}{c}{MagicBrush \cite{MagicBrush}}                                      & \multicolumn{1}{c}{\checkmark}   & \multicolumn{1}{c}{$\times$} & \multicolumn{1}{c}{$\times$}        & \multicolumn{1}{c}{10,388}         & \includegraphics[width=0.15\textwidth]{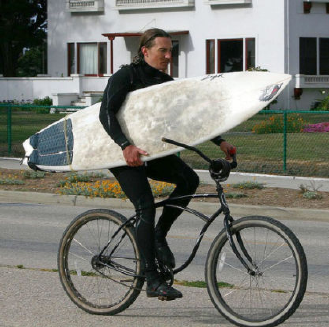}      & \includegraphics[width=0.15\textwidth]{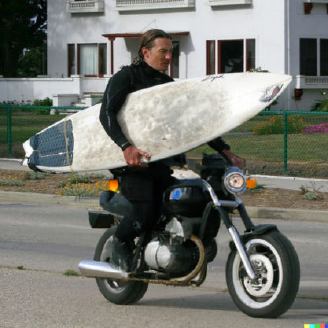}                              &  \textit{``Make the man ride a motorcycle.''}     \\ 

\multicolumn{1}{c}{InstructPix2Pix \cite{instructpix2pix}}                                      & \multicolumn{1}{c}{\checkmark}   & \multicolumn{1}{c}{\checkmark} & \multicolumn{1}{c}{$\times$}        & \multicolumn{1}{c}{\textsc{unlimited}}         & \includegraphics[width=0.15\textwidth]{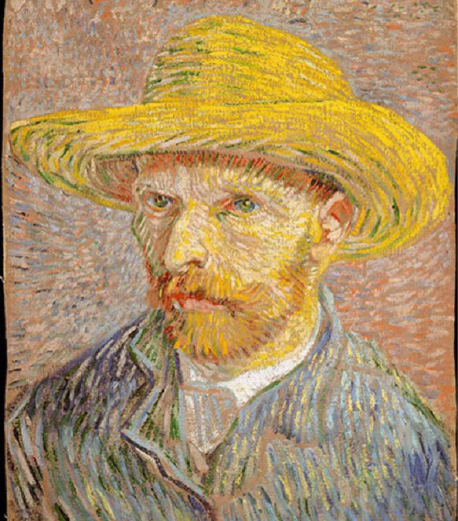}      & \includegraphics[width=0.15\textwidth]{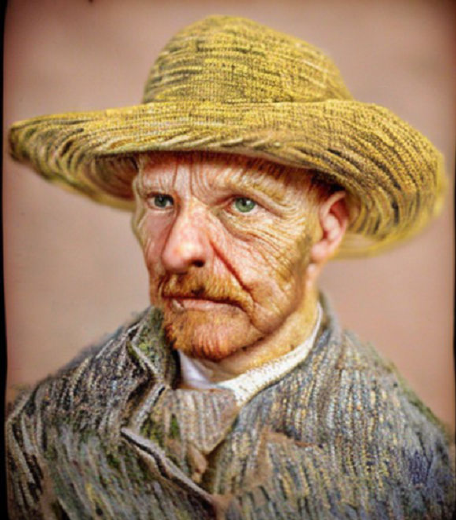}                              &  \textit{``Convert to a realistic photo.''} \\ 

\multicolumn{1}{c}{MJ-Bench \cite{mjbench}}                                      & \multicolumn{1}{c}{$\times$}   & \multicolumn{1}{c}{\checkmark} & \multicolumn{1}{c}{$\times$}        & \multicolumn{1}{c}{\textsc{unlimited}}         & \includegraphics[width=0.15\textwidth]{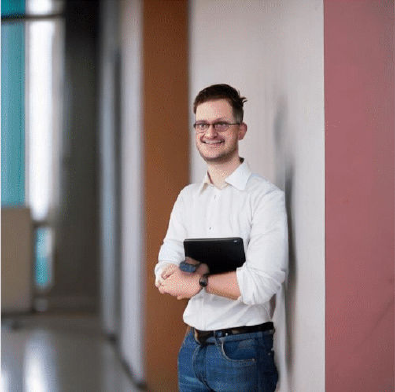}      & \includegraphics[width=0.15\textwidth]{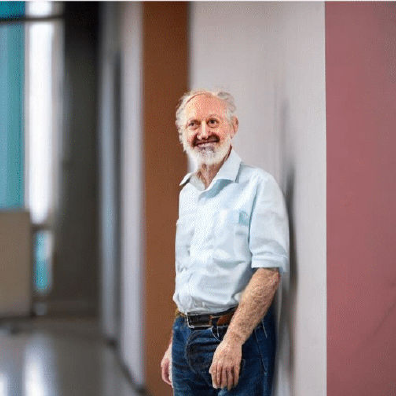}                              &  \textit{``Young or Elder. (Select)''}     \\ \midrule

\multicolumn{1}{c}{\textsc{Img-Diff}}                                 & \multicolumn{1}{c}{\checkmark}   & \multicolumn{1}{c}{\checkmark}         & \multicolumn{1}{c}{\checkmark}       & \multicolumn{1}{c}{\textsc{unlimited}}         & \includegraphics[width=0.15\textwidth]{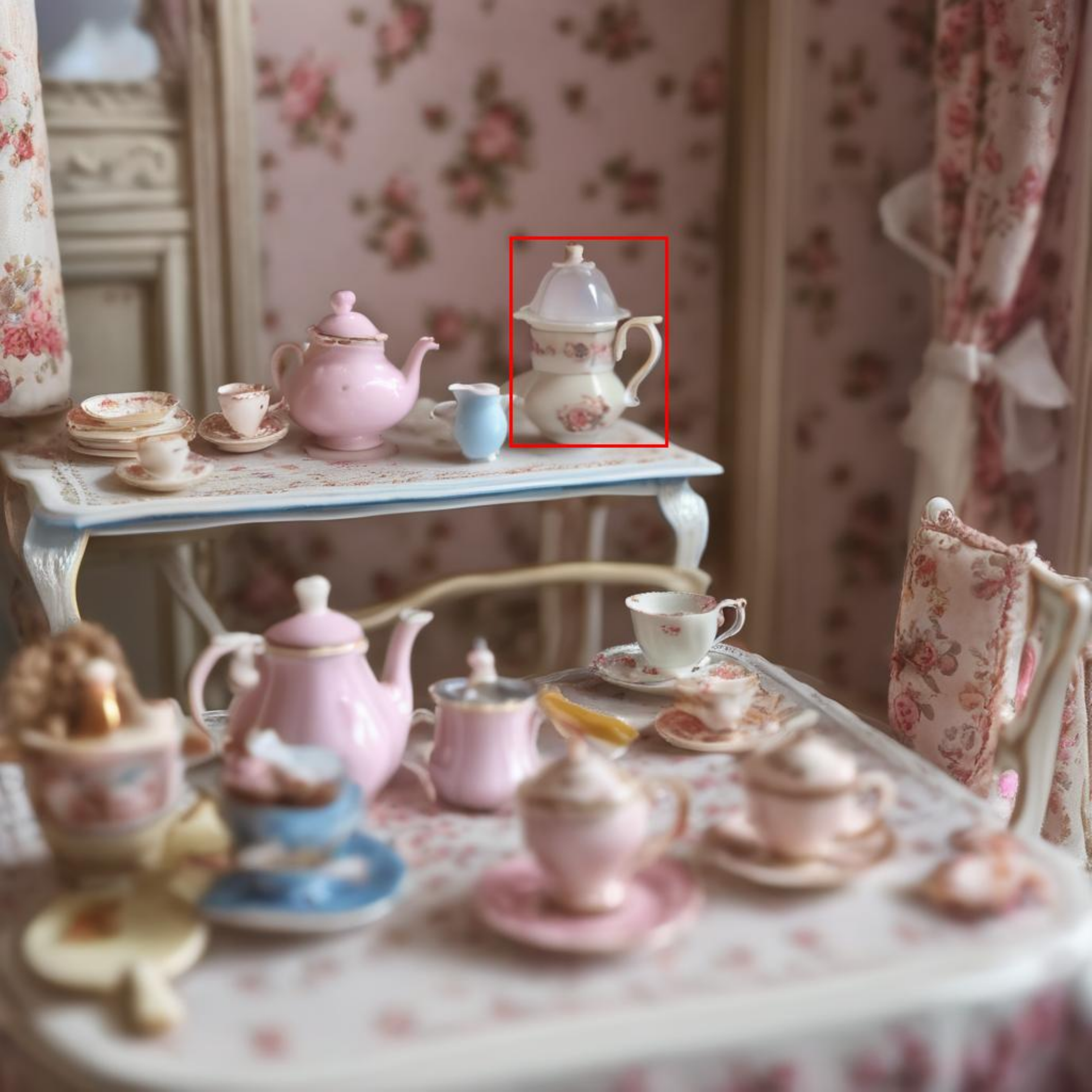}      & \includegraphics[width=0.15\textwidth]{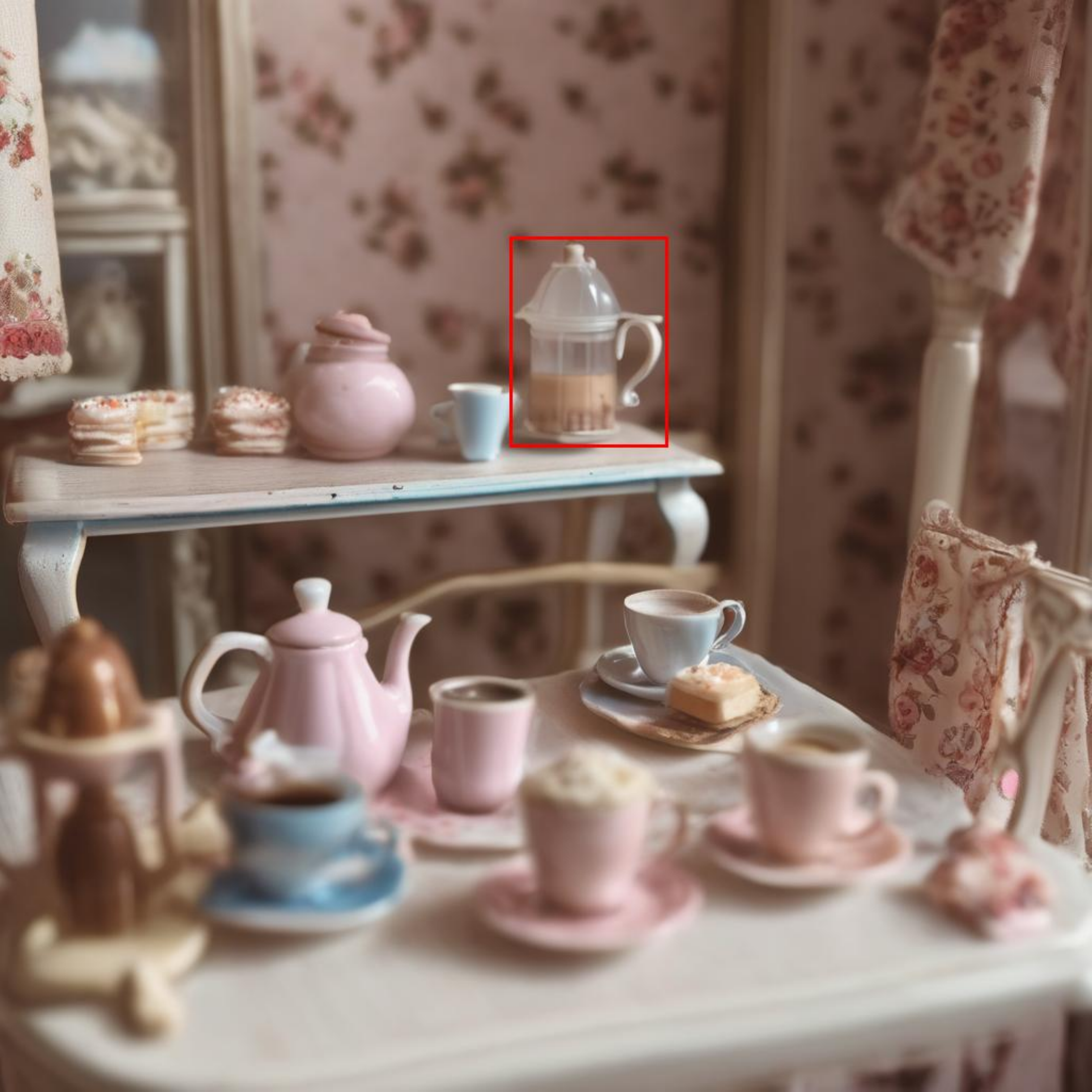}                              & \textit{``The difference is that the teapot in the right image is made of glass, whereas the teapot in the left image is made of porcelain.''}      \\ \bottomrule

\end{tabular}
}
\label{table:datasets_compare}
\end{table*}

\Cref{table:datasets_compare} compares the characteristic differences between the \textsc{Img-Diff} dataset and other existing image difference datasets. The comparison focuses on three key aspects: the ``Open-Domain'' feature, which refers to whether the dataset covers unrestricted object categories; the ``Automatic'' feature, which indicates whether the dataset can be generated fully automatically without manual intervention; and the ``Region-Focused'' feature, which highlights whether the dataset emphasizes image differences in specific detail regions rather than overall image differences.

Specifically, CUB-Birds \cite{birds-to-words} and Spot-the-Diff \cite{spot} are classic examples of traditional datasets where images are collected from the real world and data samples are generated through manual annotations. The former consists of images of various bird species captured in the wild, while the latter is compiled from street-view images taken at different time points by stationary surveillance cameras. In addition, image difference datasets can also be generated using man-made data. For example, Image-Edit-Request \cite{image-edit-request} collects image pairs consisting of manually edited images and their originals from the web, while MagicBrush \cite{MagicBrush} employs workers to write editing instructions, which are then used to generate the required image pairs with image editing techniques. These methods are limited by the scarcity of real-world data, as well as the resource and financial costs associated with manual annotation, resulting in limited dataset sizes.

To reduce resource consumption and allow for an unlimited data size, some methods have adopted fully automated generation strategies, such as InstructPix2Pix \cite{instructpix2pix} and MJ-Bench \cite{mjbench}. These methods eliminate the need for manually collected data by using generative models and image editing techniques to create image pairs. Instead of relying on human-generated annotations, they deploy high-performance VLMs or MLLMs to generate annotations. As a result, the data size is effectively limitless. However, relying on MLLMs for annotation means that these data only describe differences across the entire image. Yet, image pairs generated through image editing involve variations across multiple detailed regions. If the description only describes overall image differences, it may miss important details in fine-grained regions, resulting in inaccuracy.

Unlike the previous datasets, the \textsc{Img-Diff} dataset not only employs an automated generation pipeline but also incorporates a segmentation process to identify and capture detailed regions, which are then targeted for precise annotation. Additionally, we employ extensive filtering processes to ensure high data quality. These measures enable our dataset to achieve more comprehensive and accurate difference captions.

\subsection{Performance Comparison}
In this section, we compare the performance of the \textsc{Img-Diff} dataset with existing image difference datasets. Specifically, we incorporate each dataset separately into LLaVA’s fine-tuning data, then fine-tune LLaVA-1.5-7B and measure the performance improvements. We apply two primary dataset configurations for this comparison: the first is the CLEVR-Change \cite{duda_clevr} dataset, containing 67,600 examples. CLEVR-Change generates random 3D environments with blocks of various shapes, colors, sizes, and positions, which are subsequently altered to create image difference data. The second configuration combines the Spot-the-Diff dataset and the Image-Edit-Request dataset, totaling 13,614 samples. The results are presented in \Cref{tab:image_difference_datasets_mllm_benchmarks} and \Cref{tab:image_difference_datasets_image_difference_benchmarks}.

% \usepackage{booktabs}
% \usepackage{graphicx}
\begin{table}[h]
\centering
\caption{Performance of image difference datasets CLEVR-
Change, Image-Edit-Request \& Spot-the-Diff, and our Img-Diff dataset on MMVP and 8 MLLM benchmarks.}
\label{tab:image_difference_datasets_mllm_benchmarks}
\resizebox{\columnwidth}{!}{%
\begin{tabular}{@{}cccccc@{}}
\toprule
Model                                   & VQA$^{v2}$ & GQA       & POPE & MMB                          & MMB$^{CN}$ \\ \midrule
LLaVA-1.5-7B                            & 78.5       & 62.0      & 85.9 & 64.3                         & 58.3       \\
\textbf{LLaVA-1.5-7B + CLEVR}           & 79.2       & 63.1      & 85.7 & 65.9                         & 59.2       \\
\textbf{LLaVA-1.5-7B + ImageEdit + Spot} & 79.3       & 63.3      & 86.4 & 65.8                         & 58.9       \\
\textbf{LLaVA-1.5-7B + RP(main page)}   & 79.3       & 62.8      & 86.4 & 66.1                         & 59.8       \\ \bottomrule
                                        &            &           &      &                              &            \\ \toprule
Model                                   & MM-Vet     & SQA$^{I}$ & SEED & $\triangle$                  & MMVP       \\ \midrule
LLaVA-1.5-7B                            & 30.5       & 66.8      & 58.6 & \multicolumn{1}{c|}{-}       & 24.0       \\
\textbf{LLaVA-1.5-7B + CLEVR}           & 29.8       & 68.0      & 61.2 & \multicolumn{1}{c|}{+1.30\%} & 28.7       \\
\textbf{LLaVA-1.5-7B + ImageEdit + Spot} & 30.5       & 68.3      & 61.9 & \multicolumn{1}{c|}{+1.87\%} & 25.3       \\
\textbf{LLaVA-1.5-7B + RP(main page)}   & 33.2       & 68.2      & 61.7 & \multicolumn{1}{c|}{+3.06\%} & 27.3       \\ \bottomrule
\end{tabular}%
}
\end{table}

% \usepackage{booktabs}
% \usepackage{multirow}
% \usepackage{graphicx}
\begin{table}[h]
\centering
\caption{Performance of image difference datasets CLEVR-
Change, Image-Edit-Request \& Spot-the-Diff, and our Img-Diff dataset on image difference benchmarks.}
\label{tab:image_difference_datasets_image_difference_benchmarks}
\resizebox{\columnwidth}{!}{%
\begin{tabular}{@{}ccccc@{}}
\toprule
\multirow{2}{*}{Model}                   & \multicolumn{4}{c}{Spot-the-Diff}                                         \\ \cmidrule(l){2-5} 
                                         & BLEU             & METEOR           & CIDEr-D          & ROUGE-L          \\ \midrule
LLaVA-1.5-7B                             & 8.5              & 12.0             & 38.3             & 30.1             \\
\textbf{LLaVA-1.5-7B + CLEVR}            & \underline{9.3}  & 12.3             & \textbf{45.2}    & 30.2             \\
\textbf{LLaVA-1.5-7B + ImageEdit + Spot} & 9.1              & \underline{12.9} & 40.8             & \underline{30.5} \\
\textbf{LLaVA-1.5-7B + RP(main page)}    & \textbf{9.7}     & \textbf{13.0}    & \underline{43.2} & \textbf{30.8}    \\ \bottomrule
                                         &                  &                  &                  &                  \\ \toprule
\multirow{2}{*}{Model}                   & \multicolumn{4}{c}{Image-Edit-Request}                                    \\ \cmidrule(l){2-5} 
                                         & BLEU             & METEOR           & CIDEr-D          & ROUGE-L          \\ \midrule
LLaVA-1.5-7B                             & \underline{15.1} & 17.8             & \underline{60.6} & \underline{45.2} \\
\textbf{LLaVA-1.5-7B + CLEVR}            & \underline{15.1} & 17.8             & \textbf{60.9}    & \underline{45.2} \\
\textbf{LLaVA-1.5-7B + ImageEdit + Spot} & 13.0             & \underline{18.4} & 56.6             & 44.7             \\
\textbf{LLaVA-1.5-7B + RP(main page)}    & \textbf{16.2}    & \textbf{19.5}    & \textbf{60.9}    & \textbf{46.7}  \\ \bottomrule 
\end{tabular}%
}
\end{table}

The tables show that incorporating the CLEVR-Change, Image-Edit-Request, and Spot-the-Diff datasets into the fine-tuning of LLaVA-1.5-7B leads to performance improvements on MLLM benchmarks and image difference benchmarks. However, the performance boost from the \textsc{Img-Diff} dataset is more substantial. This could be attributed to the fact that our dataset's text is specifically generated in the format of instruction-following tasks, which provides a greater benefit for MLLMs. Additionally, our dataset places more emphasis on the image differences in detailed regions, which enhances the model's ability to capture fine-grained details, thereby improving its overall VQA capabilities and image difference recognition performance more effectively.

\section{Prioritizing Quality Over Quantity}
\label{sec:appendix_data_size}

\subsection{Discussion on Data Quantity for MLLMs}
The quality of data is generally more important than its quantity in the domain of MLLMs. As demonstrated by LLaVA-1.5, it uses only a small subset of InstructBLIP's data \cite{InstructBLIP}, supplemented with a few small-sized VQA datasets (pre-training data reduced from 129M to 558K, fine-tuning data reduced from 1.2M to 665K), achieving impressive performance and significantly surpassing those of InstructBLIP. Furthermore, a series of MLLM studies \cite{idefics2, llavanext, mgm, internvl2} validate that enhancing MLLMs' performance requires high-quality task-oriented data rather than merely increasing the volume of data.

The data volume for testing in our paper (13K and 35K) is comparable to that of many mainstream MLLM task-specific datasets, such as AI2D (12K), DocVQA (10K), ChartQA (18K), and OKVQA (9K). Despite not incorporating a large amount of data, our dataset brings appreciable performance improvement to MLLMs with modest training costs, such as elevating MGM-7B from 40 points to 50.7 on the MMVP benchmark, in which the GPT-4V gains a score of 38.7.

Considering the marginal benefits and training costs (fine-tuning 7B MLLMs on 4 A100 would take an additional 2 hours for every extra 50K samples), we aim to pursue a dataset that is small in quantity but high in quality. 
Additionally, our paper emphasizes a synthesis method rather than the dataset itself. We can generate any amount of \textsc{Img-Diff} data as needed, as our dataset is generated automatically.

\subsection{Expanding Dataset Does Not Yield Linear Performance Gains}

In addition to the 13K ``object replacement'' samples generated using MSCOCO captions on the main page, we also apply the same process and filtering thresholds to generate 34,583 samples using the captions from the LLaVA pre-training dataset. We compare the MLLM fine-tuned with the 13K samples to the one fine-tuned with the current fourfold larger dataset, aiming to explore the mathematical relationship between dataset expansion and model performance gains. The results are shown in \Cref{tab:appendix_more_data}.

% % Please add the following required packages to your document preamble:
% % \usepackage{booktabs}
% % \usepackage{graphicx}
% \begin{table*}[h]
% \centering
% \caption{Performance comparison on MMVP and 8 MLLM benchmarks (including 35K ``object replacement'' samples)}
% \label{tab:appendix_more_data}
% \resizebox{0.7\textwidth}{!}{%
% \begin{tabular}{@{}cccccccccc|c@{}}
% \toprule
% Model                                 & VQA$^{v2}$ & GQA  & POPE & MMB  & MMB$^{CN}$ & MM-Vet & SQA$^I$ & SEED & $\triangle$ & MMVP \\ \midrule
% LLaVA-1.5-7B                          & 78.5       & 62.0 & 85.9 & 64.3 & 58.3       & 30.5   & 66.8    & 58.6 & - & 24.0                           \\
% \textbf{LLaVA-1.5-7B+RP(13K)}         & 79.3       & 62.8 & 86.4 & 66.1 & 59.8       & 33.2   & 68.2    & 61.7 & +3.06\% & 27.3                     \\
% \textbf{LLaVA-1.5-7B+RP(13K)+RP(35K)} & 79.2       & 63.1 & 86.2 & 66.9 & 59.2       & 33.3   & 69.0    & 62.2 & +3.40\%  & 31.3                    \\ \bottomrule
% \end{tabular}%
% }
% \end{table*}

% \usepackage{booktabs}
% \usepackage{graphicx}
\begin{table}[h]
\centering
\caption{Performance comparison on MMVP and 8 MLLM benchmarks (including 35K ``object replacement'' samples).}
\label{tab:appendix_more_data}
\resizebox{\columnwidth}{!}{%
\begin{tabular}{@{}cccccc@{}}
\toprule
Model                                 & VQA$^{v2}$           & GQA                  & POPE                 & MMB                              & MMB$^{CN}$           \\ \midrule
LLaVA-1.5-7B                          & 78.5                 & 62.0                 & 85.9                 & 64.3                             & 58.3                 \\
\textbf{LLaVA-1.5-7B + RP(13K)}         & 79.3                 & 62.8                 & 86.4                 & 66.1                             & 59.8                 \\
\textbf{LLaVA-1.5-7B + RP(13K) + RP(35K)} & 79.2                 & 63.1                 & 86.2                 & 66.9                             & 59.2                 \\ \bottomrule
\multicolumn{1}{l}{}                  & \multicolumn{1}{l}{} & \multicolumn{1}{l}{} & \multicolumn{1}{l}{} & \multicolumn{1}{l}{}             & \multicolumn{1}{l}{} \\ \toprule
Model                                 & MM-Vet               & SQA$^I$              & SEED                 & $\triangle$ & MMVP                 \\ \midrule
LLaVA-1.5-7B                          & 30.5                 & 66.8                 & 58.6                 & \multicolumn{1}{c|}{-}           & 24.0                 \\
\textbf{LLaVA-1.5-7B + RP(13K)}         & 33.2                 & 68.2                 & 61.7                 & \multicolumn{1}{c|}{+3.06\%}     & 27.3                 \\
\textbf{LLaVA-1.5-7B + RP(13K) + RP(35K)} & 33.3                 & 69.0                 & 62.2                 & \multicolumn{1}{c|}{+3.40\%}     & 31.3                 \\ \bottomrule
\end{tabular}%
}
\end{table}

% \usepackage{booktabs}
% \usepackage{multirow}
% \usepackage{graphicx}
\begin{table}[h]
\centering
\caption{Performance comparison on Spot-the-Diff and Image-Edit-Request (including 35K ``object replacement'' samples).}
\label{tab:appendix_more_data_imgdiff_benchmarks}
\resizebox{\columnwidth}{!}{%
\begin{tabular}{@{}ccccc@{}}
\toprule
\multirow{2}{*}{Model}                  & \multicolumn{4}{c}{Spot-the-Diff}                                         \\ \cmidrule(l){2-5} 
                                       & BLEU             & METEOR           & CIDEr-D          & ROUGE-L          \\ \midrule
LLaVA-1.5-7B                            & 8.5              & 12.0             & 38.3             & 30.1             \\
\textbf{LLaVA-1.5-7B + RP(13K)}          & \underline{9.7}  & \underline{13.0} & \underline{43.2} & \underline{30.8} \\
\textbf{LLaVA-1.5-7B + RP(13K) + RP(35K)} & \textbf{9.8}     & \textbf{13.1}    & \textbf{45.3}    & \textbf{31.0}    \\ \bottomrule
\multicolumn{1}{l}{}                                         & \multicolumn{1}{l}{} & \multicolumn{1}{l}{} & \multicolumn{1}{l}{} & \multicolumn{1}{l}{} \\ \toprule
\multirow{2}{*}{Model}                  & \multicolumn{4}{c}{Image-Edit-Request}                                    \\ \cmidrule(l){2-5} 
                                        & BLEU             & METEOR           & CIDEr-D          & ROUGE-L          \\ \midrule
LLaVA-1.5-7B                            & 15.1             & 17.8             & 60.6             & 45.2             \\
\textbf{LLaVA-1.5-7B + RP(13K)}          & \underline{16.2} & \textbf{19.5}    & \underline{60.9} & \underline{46.7} \\
\textbf{LLaVA-1.5-7B + RP(13K) + RP(35K)} & \textbf{16.4}    & \underline{19.1} & \textbf{65.5}    & \textbf{46.8}    \\ \bottomrule
\end{tabular}%
}
\end{table}

We observe that the average performance gain on the MLLM benchmarks has become 3.40\%, while the performance gain from the previous \textsc{Img-Diff} dataset was 3.06\%. On the MMVP benchmark, the model fine-tuned with more data achieves further improvement, raising its score from 27.3, obtained with 13K samples, to the current score of 31.3. Furthermore, on Spot-the-Diff and Image-Edit-Request, the additional data also contributes to further performance gains. These results indicate that a moderate increase in data size can further enhance model performance.

Although adding more data can improve the MLLM's performance, it is worth noting that while we quadruple the dataset, the performance improvements do not increase by a factor of four. This aligns with the fact that the relationship between the data size and performance gains is not linear. As we increase the amount of similar data, the performance gains eventually reach a maximum limit. For future work, further investigation can be conducted into the relationship between different data volumes and performance improvements under the same filtering threshold.

\section{Expanding Diversity with Lexicons}
\label{sec:appendix_diversity_lexicon}
On the main page, beyond the intrinsic diversity of object names within the caption database, we increase the temperature of the LLM used for object name substitution to enhance the randomness of model outputs. This helps us expand the range of object categories covered by our dataset. Additionally, we experiment with randomly selecting nouns from an object name lexicon to replace original object names in captions, further enriching the dataset's diversity. This section provides a detailed explanation of this ``Expanding Diversity with Lexicons'' method and the experimental results on LLaVA-1.5-7B.

To construct the object name lexicon, we initially use the NLTK tool to filter all nouns from the WordNet lexicon. Next, we categorize each word based on its synsets entries, labeling them accordingly. Finally, we select object names classified under ``machine,'' ``living\_thing,'' ``natural\_object,'' ``fruit,'' ``vehicle,'' ``container,'' ``clothing,'' ``fixture,'' ``appliance,'' ``furniture,'' or ``food'' and form the final object name lexicon. The resulting lexicon comprises 5,526 distinct object names.

Following this, as described on the main page, we generate a test dataset using MSCOCO captions. Specifically, we replace object names in MSCOCO captions randomly with nouns of the same category from the object name lexicon, forming caption pairs that are later used for further generation and filtering processes. This approach resulted in 8,930 high-quality ``object replacement'' samples. We utilize this data to fine-tune LLaVA-1.5-7B, obtaining the results shown in \Cref{tab:appendix_diversity_mllm_benchmarks}.

% % Please add the following required packages to your document preamble:
% % \usepackage{booktabs}
% % \usepackage{graphicx}
% \begin{table*}[t]
% \centering
% \caption{Performance comparison on 8 MLLM benchmarks (using data generated with lexicons.)}
% \label{tab:appendix_diversity_mllm_benchmarks}
% \resizebox{0.7\textwidth}{!}{%
% \begin{tabular}{@{}cccccccccc@{}}
% \toprule
% Model                             & VQA$^{v2}$ & GQA  & POPE & MMB  & MMB$^{CN}$ & MM-Vet & SQA$^{I}$ & SEED & $\triangle$ \\ \midrule
% LLaVA-1.5-7B                      & 78.5       & 62   & 85.9 & 64.3 & 58.3       & 30.5   & 66.8      & 58.6 & -           \\
% \textbf{LLaVA-1.5-7B+RP(main page)} & 79.3          & 62.8 & 86.4 & 66.1          & 59.8 & 33.2         & 68.2          & 61.7          & $+3.06\%$  \\
% \textbf{LLaVA-1.5-7B+RP(lexicon)} & 79.2       & 62.7 & 86.3 & 66.2 & 59.4       & 32.2   & 68.8      & 61.8 & +2.67\%     \\ \bottomrule
% \end{tabular}%
% }
% \end{table*}

% \usepackage{booktabs}
% \usepackage{graphicx}
\begin{table}[h]
\centering
\caption{Performance comparison on MMVP and 8 MLLM benchmarks (using data generated with lexicons).}
\label{tab:appendix_diversity_mllm_benchmarks}
\resizebox{\columnwidth}{!}{%
\begin{tabular}{@{}cccccc@{}}
\toprule
Model                                 & VQA$^{v2}$           & GQA                  & POPE                 & MMB                              & MMB$^{CN}$           \\ \midrule
LLaVA-1.5-7B                          & 78.5                 & 62                   & 85.9                 & 64.3                             & 58.3                 \\
\textbf{LLaVA-1.5-7B + RP(main page)} & 79.3                 & 62.8                 & 86.4                 & 66.1                             & 59.8                 \\
\textbf{LLaVA-1.5-7B + RP(lexicon)}   & 79.2                 & 62.7                 & 86.3                 & 66.2                             & 59.4                 \\ \bottomrule
\multicolumn{1}{l}{}                  & \multicolumn{1}{l}{} & \multicolumn{1}{l}{} & \multicolumn{1}{l}{} & \multicolumn{1}{l}{}             & \multicolumn{1}{l}{} \\ \toprule
Model                                 & MM-Vet               & SQA$^{I}$            & SEED                 & $\triangle$ & MMVP                 \\ \midrule
LLaVA-1.5-7B                          & 30.5                 & 66.8                 & 58.6                 & \multicolumn{1}{c|}{-}           & \multicolumn{1}{|c}{24.0} \\
\textbf{LLaVA-1.5-7B + RP(main page)} & 33.2                 & 68.2                 & 61.7                 & \multicolumn{1}{c|}{$+3.06\%$}   & \multicolumn{1}{|c}{27.3} \\
\textbf{LLaVA-1.5-7B + RP(lexicon)}   & 32.2                 & 68.8                 & 61.8                 & \multicolumn{1}{c|}{+2.67\%}     & \multicolumn{1}{|c}{30.0} \\ \bottomrule
\end{tabular}%
}
\end{table}

% \usepackage{booktabs}
% \usepackage{multirow}
% \usepackage{graphicx}
\begin{table}[h]
\centering
\caption{Performance comparison on Spot-the-Diff and Image-Edit-Request(using data generated with lexicons).}
\label{tab:appendix_diversity_diff_benchmarks}
\resizebox{\columnwidth}{!}{%
\begin{tabular}{@{}ccccc@{}}
\toprule
\multirow{2}{*}{Model}     & \multicolumn{4}{c}{Spot-the-Diff}      \\ \cmidrule(l){2-5} 
                           & BLEU   & METEOR  & CIDEr-D  & ROUGE-L  \\ \midrule
LLaVA-1.5-7B               & 8.5    & 12.0    & 38.3     & \underline{30.1}     \\
\textbf{LLaVA-1.5-7B + RP(main page)}          & \textbf{9.7}  & \textbf{13.0} & \textbf{43.2} & \textbf{30.8} \\
\textbf{LLaVA-1.5-7B + RP(lexicon)} & \underline{8.9}    & \underline{12.2}    & \underline{41.9}     & 29.9     \\ \bottomrule
                           &        &         &          &          \\ \toprule
\multirow{2}{*}{Model}     & \multicolumn{4}{c}{Image-Edit-Request} \\ \cmidrule(l){2-5} 
                           & BLEU   & METEOR  & CIDEr-D  & ROUGE-L  \\ \midrule
LLaVA-1.5-7B               & \underline{15.1}   & 17.8    & \underline{60.6}     & 45.2     \\
\textbf{LLaVA-1.5-7B + RP(main page)}          & \textbf{16.2} & \textbf{19.5}    & \textbf{60.9} & \underline{46.7} \\
\textbf{LLaVA-1.5-7B + RP(lexicon)} & 13.9   & \underline{19.4}    & 60.4     & \textbf{46.9}     \\ \bottomrule
\end{tabular}%
}
\end{table}

As shown in \Cref{tab:appendix_diversity_mllm_benchmarks} and \Cref{tab:appendix_diversity_diff_benchmarks}, the current dataset still provides significant performance improvements for LLaVA-1.5-7B. Specifically, the fine-tuned MLLM achieves comprehensive performance improvement across eight MLLM benchmarks, with improvement levels comparable to those on the main page, resulting in an average performance increase of 2.67\%. Besides, the current dataset also improves the performance of LLaVA-1.5-7B on image difference benchmarks.

By using a lexicon for object name replacement, we can more effectively enhance the diversity of the \textsc{Img-Diff} dataset. Specifically, we can increase the number of noun samples included in the lexicon, as well as perform multiple rounds of noun replacement on the same caption. As a result, the quality of our data can be further improved.

\section{Performance Based on Contrastive Chain-of-Thought}
\label{appendix_cocot}
In addition to the standard VQA evaluation, we also assess the \textsc{Img-Diff} dataset using the Contrastive Chain-of-Thought (CoCoT \cite{zhang2024cocot}) method. This evaluation method involves prompting the model with the instruction, ``Please identify the similarities and differences between these two images,'' and requiring the MLLM to pinpoint the differences before it answers the final VQA question. The differences identified are then used as context-enhanced text to support its own response to the VQA task.

% \usepackage{multirow}
% \usepackage{graphicx}
\begin{table}[h]
\centering
\caption{Results on MMVP Using the CoCoT Method.}
\label{tab:appendix_cocot}
\resizebox{0.6\columnwidth}{!}{%
\begin{tabular}{ccc}
\toprule
\multirow{2}{*}{Model}     & \multicolumn{2}{c}{MMVP}                                     \\ \cmidrule(l){2-3}
                           & \multicolumn{1}{l}{w/ CoCot} & \multicolumn{1}{l}{w/o CoCot} \\ \midrule
LLaVA-1.5-7B               & 24.0                         & 22.0                          \\
\textbf{LLaVA-1.5-7B + RP} & 27.3                         & 29.0                          \\ \bottomrule
\end{tabular}%
}
\end{table}

We test the original LLaVA-1.5-7B and our fine-tuned model on the MMVP benchmark using CoCoT. As shown in \Cref{tab:appendix_cocot}, the original model's score drops from 24 to 22, while the score of the model fine-tuned with our data rises from 27.3 to 29. This indicates that, after fine-tuning without \textsc{Img-Diff} data, the MLLM demonstrates an enhanced ability to recognize image differences and can generate more accurate descriptive information to support VQA tasks.

\section{Testing on MLLMs at Different Scales}
\label{sec:appendix_other_mllms}

On the main page, we primarily conduct experiments on MLLMs with a 7B scale. In this section, we will explore the impact of our dataset on models of different sizes. Specifically, we fine-tune LLaVA-1.5-13B and InternVL2-1B, representing a larger and a smaller model. We then test these models on both MLLM benchmarks and image difference benchmarks.

% \usepackage{booktabs}
% \usepackage{graphicx}
\begin{table}[h]
\centering
\caption{Performance of ``object replacement'' data on LLaVA-1.5-13B and InternVL2-1B (evaluations on MMVP and 8 MLLM Benchmarks).}
\label{tab:appendix_other_mllm_mllm_benmarks}
\resizebox{\columnwidth}{!}{%
\begin{tabular}{@{}cccccc@{}}
\toprule
Model                       & VQA$^{v2}$ & GQA       & POPE & MMB                          & MMB$^{CN}$ \\ \midrule
LLaVA-1.5-13B               & 80.0         & 63.3      & 85.9 & 67.7                         & 63.6       \\
\textbf{LLaVA-1.5-13B + RP} & 80.3       & 64.1      & 86.6 & 69.2                         & 63.2       \\ \midrule
InternVL2-1B-FT             & 77.3       & 60.2      & 86.6 & 68.6                         & 60.7       \\
\textbf{InternVL2-1B + RP}  & 77.4       & 60.2      & 87.1 & 69.0                           & 60.7       \\ \bottomrule
                            &            &           &      &                              &            \\ \toprule
Model                       & MM-Vet     & SQA$^{I}$ & SEED & $\triangle$                  & MMVP       \\ \midrule
LLaVA-1.5-13B               & 35.4       & 71.6      & 61.6 & \multicolumn{1}{c|}{-}       & 24.7       \\
\textbf{LLaVA-1.5-13B + RP} & 37.4       & 71.7      & 62.9 & \multicolumn{1}{c|}{+1.49\%} & 32.0       \\ \midrule
InternVL2-1B-FT             & 31.9       & 88.5      & 61.4 & \multicolumn{1}{c|}{-}       & 16.0       \\
\textbf{InternVL2-1B + RP}  & 33.4       & 88.7      & 61.7 & \multicolumn{1}{c|}{+0.84\%}                      & 18.0       \\ \bottomrule
\end{tabular}%
}
\end{table}
% \usepackage{multirow}
% \usepackage{graphicx}
% \usepackage[normalem]{ulem}
\begin{table}[h]
\centering
\caption{Performance of ``object replacement'' data on LLaVA-1.5-13B and InternVL2-1B (evaluations on Spot-the-Diff and Image-Edit-Request).}
\label{tab:appendix_other_mllm_image_diff_benmarks}
\resizebox{0.95\columnwidth}{!}{%
\begin{tabular}{ccccc}
\toprule
\multirow{2}{*}{Model}      & \multicolumn{4}{c}{Spot-the-Diff}                             \\ \cmidrule(l){2-5} 
                            & BLEU          & METEOR        & CIDEr-D       & ROUGE-L       \\ \midrule
LLaVA-1.5-13B               & {\underline{9.7}}     & {\underline{12.3}}    & {\underline{44.6}}    & {\underline{31.0}}    \\
\textbf{llava-1.5-13b + RP} & \textbf{9.9}  & \textbf{13.1} & \textbf{45.8} & \textbf{31.4} \\ \midrule
InternVl2-1B-FT             & 6.5           & 11.4          & 24.7          & 26.5          \\
\textbf{InternVl2-1B + RP}  & 6.9           & 11.5          & 25.7          & 26.5          \\ \bottomrule
                            &               &               &               &               \\ \toprule
\multirow{2}{*}{Model}      & \multicolumn{4}{c}{Image-Edit-Request}                        \\ \cmidrule(l){2-5} 
                            & BLEU          & METEOR        & CIDEr-D       & ROUGE-L       \\ \midrule
LLaVA-1.5-13B               & \textbf{16.6} & {\underline{18.0}}    & {\underline{62.9}}    & {\underline{46.2}}    \\
\textbf{llava-1.5-13b + RP} & {\underline{15.9}}    & \textbf{20.1} & \textbf{65.3} & \textbf{47.2} \\ \midrule
InternVl2-1B-FT             & 7.3           & 11.6          & 28.7          & 35.3          \\
\textbf{InternVl2-1B + RP}  & 8.4           & 12.1          & 30.3          & 37.4          \\ \bottomrule
\end{tabular}%
}
\end{table}

\Cref{tab:appendix_other_mllm_mllm_benmarks} and \Cref{tab:appendix_other_mllm_image_diff_benmarks} show that our dataset remains effective on LLaVA-1.5-13B and InternVL2-1B, delivering comprehensive performance improvements across eight MLLM benchmarks and the image difference benchmarks. This demonstrates the versatility of our dataset, proving its capability to enhance model performance not only for 7B-scale models but also for smaller or larger models.

\section{Ablation Studies}
\label{sec:ablation_study}
% \usepackage{booktabs}
% \usepackage{graphicx}
\begin{table*}[t]
\centering
\caption{The impact of different filtering thresholds on the performance of our dataset.}
\label{tab:ablation}
\resizebox{0.8\textwidth}{!}{%
\begin{tabular}{@{}lccccccccc@{}}
\toprule
\multicolumn{1}{c}{Threshold}                             & VQA$^{v2}$ & GQA  & POPE & MMB & MMB$^{CN}$ & MM-Vet & SQA${^I}$ & SEED & $\triangle$ \\ \midrule
\multicolumn{1}{c}{LLaVA-1.5-7B}                                              & 78.5  & 62.0 & 85.9 & 64.3    & 58.3       & 30.5   & 66.8      & 58.6 & -       \\
\textbf{(1) IS 0.9-0.98 + BITM 0.3 + CS   0.9 + CITM 0.3}      & 79.1  & 62.3 & 86.0 & 66.8    & 59.5       & 32.7   & 66.6      & 61.6 & $+2.42\%$   \\
\textbf{(2) IS 0.9-0.98 + BITM 0.35 + CS   0.9 + CITM 0.3}     & 79.1  & 62.2 & 85.9 & 66.7    & 59.5       & 32.7   & 67.1      & 61.9 & $+2.52\%$   \\
\textbf{(3) IS 0.9-0.98 + BITM 0.35 + CS   0.85 + CITM 0.4}    & 79.3  & 62.8 & 86.4 & 66.1    & 59.8       & 33.2   & 68.2      & 61.7 & $+3.06\%$   \\
\textbf{(4) IS 0.85-0.98 + BITM 0.35 + CS   0.85 + CITM 0.4}   & 79.2  & 62.7 & 86.3 & 66.2    & 57.4       & 32.2   & 68.8      & 61.8 & $+2.24\%$   \\
\bottomrule
\end{tabular}%
}
\end{table*}

To investigate the impact of filtering thresholds on our data performance, we set different filtering thresholds and generate various versions of our ``object replacement'' dataset. We then finetune multiple versions of LLaVA-1.5-7B using these datasets and evaluate their performance on commonly used MLLM benchmarks. Specifically, the threshold for the Image Similarity Filter of the Difference Area Generator is abbreviated as \textbf{IS} (Image Similarity). The threshold for the Image-Text Matching Filter of the Difference Area Generator is abbreviated as \textbf{BITM} (Bounding Box Image-Text Matching). The threshold for the Caption Similarity Filter of the Difference Captions Generator is abbreviated as \textbf{CS} (Captions Similarity). The threshold for the Image-Text Matching Filter of the Difference Captions Generator is abbreviated as \textbf{CITM} (Captions Image-Text Matching). The evaluation results are shown in \Cref{tab:ablation}.

\paragraph{Image Similarity (IS)} Based on \Cref{tab:ablation}, Model (3) adjusts the IS threshold from 0.9-0.98 to 0.85-0.98 compared to Model (4), reducing the filtering intensity for the similarity of image pairs. This adjustment leads to a significant performance decline, indicating that the similarity of image pairs has a substantial impact on data quality. When the similarity is low, the data generation process may introduce more ineffective instances, as segmentation could generate more areas unrelated to the valid objects (i.e., the replaced or replacing objects).

\paragraph{Bounding Box Image-Text Matching (BITM)} Model (2), compared to Model (1), increases the BITM threshold, meaning that when filtering to obtain valid bounding boxes, only those more likely to contain valid objects are retained. After raising the threshold, slight improvements in model performance are observed, which demonstrates that only bounding boxes more related to the replaced or replacing objects should be retained.

\paragraph{Captions Similarity (CS) and Captions Image-Text Matching (CITM)} Model (3) increases both the CS threshold and the CITM threshold compared to Model (2). Raising the CS threshold implies a greater filtering strength for similar captions, which means that if the two objects corresponding to the same bounding box coordinate in an image pair are similar, the bounding box will be filtered out. As for the CITM threshold, increasing the CITM threshold aims to enhance the alignment between the captions and the objects being described. After raising both the CS and CITM thresholds, the model's performance shows a significant improvement.

Based on \Cref{tab:ablation}, it can be concluded that the stronger the filtering intensity, the better our dataset's effectiveness. However, due to the increased filtering intensity resulting in a reduced number of final instances, we choose the settings of Model (3) as our optimal threshold to ensure a sufficient number of generated instances. In our future work, we will expand the data sources to generate more pairs of similar images and then evaluate the effects of data obtained with higher filtering intensity.

\section{Additional Details of Experiments }
\label{appendix:additional_details}
\subsection{Preprocessing of image pairs before inputting into MLLMs during training and inference}
The MLLMs selected in our paper (LLaVA-1.5, MGM, InternVL2) only support single-image input. Therefore, our image pairs need to be horizontally concatenated before being fed into MLLMs' image encoder.
Specifically, we horizontally concatenate the images in pairs and add a vertical black dividing line, 20 pixels wide, between the images.

\subsection{Training Process for MLLMs}
The training process for advanced MLLMs, including LLaVA-1.5, MGM and InternVL2, typically involves two stages: the pre-training stage and fine-tuning stage. During the pre-training stage, the MLLMs keep the backbone LLM and the vision encoder frozen and zero-initialize the learnable projector which is used for semantic mapping and cross-modality alignment. Only the projector is trained using the pre-training dataset. In the fine-tuning stage, we unfreeze the backbone LLM and fine-tune both the backbone LLM and the learnable projector using the visual instruction tuning dataset. Specifically, the pre-training dataset is usually an image captioning dataset, while the visual instruction tuning dataset typically consists of VQA datasets for various tasks. Thus, the \textsc{Img-Diff} dataset is integrated into the visual instruction tuning dataset during the fine-tuning stage and used together with the original dataset to fine-tune the MLLMs.

\subsection{Model Selection}
\label{appendix:model_selection}
\subsubsection{Overview}
The models used in our project are among the best-performing ones identified for the tasks assigned to them. Besides, they are interchangeable. Therefore, if better model options become available, researchers can replace the current models with those that offer superior performance to achieve a more effective dataset.

\subsubsection{Selection of the Semantic Segmentation Model}
In our project, we need to use a semantic segmentation model to identify regions containing objects in images. To ensure a diverse range of object categories is covered, we opt for models like SAM \cite{sam} instead of traditional semantic segmentation models. Furthermore, to reduce time consumption, we select FastSAM, one of the most efficient and effective models within the SAM-like category, as our segmentation model.

\subsubsection{Model Size}
Considering the device limitation and time consumption, our paper utilizes the LLM Vicuna-1.5-13B \cite{vicuna} for object name replacement in the image pairs generation process. For semantic segmentation in the Difference Area Generator, the FastSAM-x model is employed. For the CLIP model, we choose ``clip-vit-base-patch32'', and for the BLIP model, we select ``blip-itm-large-coco''. In the Difference Captions Generator, we use the MLLM LLaVA-NEXT-13B to generate content captions and difference captions. These models are interchangeable. When resources allow, researchers can substitute them with higher-performance models to achieve datasets with improved performance.

\subsection{Filtering Thresholds}
\label{appendix:filtering_thres}
During the generation process of ``object replacement'' data, we employ multiple filtering operations. In this subsection, we will outline the filtering thresholds we use.

In the Difference Area Generator, we use FastSAM to perform semantic segmentation on images and obtain bounding box information for regions where objects might be present. To ensure we gather a sufficient number of candidate regions, we set the confidence score threshold to 0.05, which means that we consider a region to contain objects when its confidence score is greater than 0.05. Additionally, to prevent overlapping regions, we set the Intersection over Union (IoU) threshold to 0.5.

At the beginning stage of the Difference Area Generator, before using FastSAM for segmentation, we employ the Image Similarity Filter to retain only those with similarity between 0.9 and 0.98. This ensures that the image pairs are highly similar but not identical.

In the Difference Detector stage of the Difference Area Generator, after cropping sub-images based on the bounding box information, we use the Image Similarity Filter to filter the sub-image pairs and consider them to be different only when the similarity score is less than 0.85.

In the mid-stage of the Difference Area Generator, after performing sub-image cropping based on the bounding box information, we use the Image-text Matching Filter to determine whether these sub-images contain valid objects. When the score exceeds 0.35, we consider the sub-image to contain valid objects, and the bounding box is deemed effective.

In the Difference Area Generator, after obtaining all effective bounding boxes, we use the IoU method to filter out the overlapping ones. We set the IoU threshold to 0.5, retaining only the bounding boxes with a higher degree of difference for similar positions.

In stage 1 of the Difference Captions Generator, after cropping the images into sub-images and generating content captions, we use the Image-text Matching Filter to evaluate the matching degree between the sub-images and the captions. We only consider a caption to be correct if the image-text matching score exceeds 0.4.

In stage 1 of the Difference Captions Generator, we use the Captions Similarity Filter to determine whether the two content captions of an image pair, describing the regions of the same bounding box, are different. We use CLIP to obtain text features for the two captions and then calculate the cosine similarity between them. When the cosine similarity is below 0.85, we consider the two captions to be different.

Setting the filtering intensity too high may lead to a reduced number of remaining samples. To ensure that the dataset still has enough samples after filtering, we outline adjustable thresholds as described above. As mentioned in \Cref{sec:ablation_study}, higher filtering intensity typically results in better model performance. Therefore, researchers may consider expanding the data sources and increasing the filtering intensity to improve dataset performance.

\subsection{Resource and Time Consumption}
With four NVIDIA A100 GPUs, it took 4.5 days to synthesize 118K high-quality image pairs. The subsequent filtering and description-generating processes took approximately two days in total.

\begin{figure*}[t]
\begin{center}
\centerline{\includegraphics[width=\textwidth]{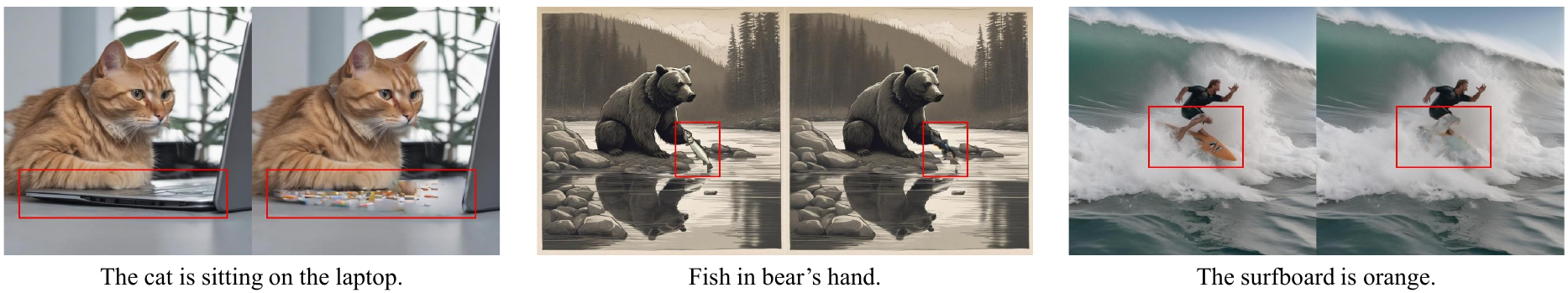}}
\caption{Three ``object removal'' examples.}
\label{fig:object_removal_example}
\end{center}
\end{figure*}

\begin{figure*}[t]
\begin{center}
\centerline{\includegraphics[width=0.8\textwidth]{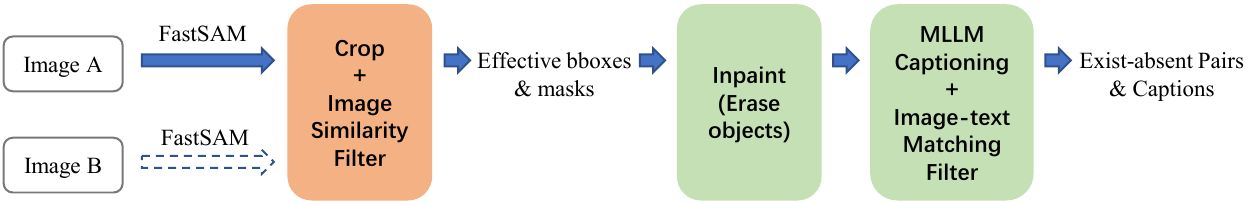}}
\caption{An overview of the generation steps for ``object removal'' data.}
\label{fig:exist_absent_pairs_generator}
\end{center}
\end{figure*}

\section{The ``Object Removal'' Exploration}
\label{appendix_rm_data}
\subsection{Overview}
On the main page, we generate pairs of similar images focusing on object replacement. Their bounding box regions generally contain objects. However, the ability to determine the object's presence is also crucial. Thus, we generate another set of image pairs where the difference lies in the presence or absence of objects, to enhance the model's ability to determine object presence. We refer to these image pairs as ``exist-absent pairs'' and the data as ``object removal'' data.

\subsection{Generation Process}
\subsubsection{Workflow}
``Object removal'' involves erasing a specific object from an image and then merging the edited image with the original to form an exist-absent pair. The detailed workflow is as follows: first, FastSAM is used to segment the image, which provides a set of bounding boxes and masks. Next, an Image Similarity Filter is applied to filter the bounding boxes and accompanying masks, keeping only those that contain objects. Then, we use the text-to-image generative model SDXL-turbo\cite{sdxl-turbo} to inpaint the images with the remaining masks, erasing specific objects from the images and generating exist-absent pairs. Next, we use an MLLM to describe the removed object for each exist-absent pair, and a filter is employed to verify the accuracy of the description. Finally, we draw red boxes on images based on the bounding box information, and then the object descriptions are converted into multiple-choice questions, such as: ``\textit{which image has the object related to `DESCRIPTION' within the red bounding box? A. the left image B. the right image.}'' Here, \textit{DESCRIPTION} refers to the description of the erased objects. After all processing and filtering, we obtain 5,773 pieces of ``object removal'' data. The general framework is shown in \Cref{fig:exist_absent_pairs_generator}.

\subsubsection{Image Similarity Filter}
In the current process, the function of the Image Similarity Filter is to filter out the bounding box regions that do not contain objects. For each image, we need its corresponding image in the image pair generated in \Cref{sec:image_pairs_generation} to determine whether its bounding box regions contain objects. Since the image pairs are generated by replacing objects, the difference areas between the two images are highly likely to be the regions containing valid objects. Therefore, for each bounding box, we crop the sub-images from image A (the current image) and image B (the other image in the pair), and then calculate the similarity of these two sub-images. When the similarity is below 0.9, we consider these two sub-images to be different, indicating that the bounding box region contains an object.

\subsubsection{Erase Objects}
We use the generative model SDXL-turbo\cite{sdxl-turbo} to erase objects based on the masks obtained during segmentation. The prompt is ``background, nothing, 8k.'' After inpainting, the object in the masked regions is erased, while the rest of the image remains unchanged. Hence, we obtain exist-absent pairs.

\subsubsection{MLLM Captioning}
We use the MLLM LLaVA-NEXT to generate descriptions for the erased objects. Specifically, we provide the MLLM with the bounding box coordinates and ask it to describe the corresponding area in the original image. Subsequently, we crop the exist-absent pairs based on the bounding box information and then use an Image-Text Matching Filter to assess the matching degree between the sub-images and the descriptions. If the matching score between the sub-image containing objects and its description is greater than 0.35, and the matching score between the sub-image not containing objects and its description is less than 0.2, we consider the description to be accurate and the exist-absent pair to be valid.

% % Please add the following required packages to your document preamble:
% % \usepackage{booktabs}
% % \usepackage{graphicx}
% % \usepackage[normalem]{ulem}
% % \useunder{\uline}{\ul}{}
% \begin{table*}[t]
% \centering
% \caption{Performance comparison on 8 MLLM benchmarks (including ``object removal'' data).}
% \label{tab:rm_mllm_benchmark}
% \resizebox{0.6\textwidth}{!}{%
% \begin{tabular}{@{}cccccccccc@{}}
% \toprule
% Model                   & VQA$^{v2}$         & GQA           & POPE          & MMB       & MMB$^{CN}$    & MM-Vet        & SQA$^{I}$     & SEED          & $\triangle$ \\ \midrule
% LLaVA-7B                & 78.5          & 62.0          & 85.9          & 64.3          & 58.3          & 30.5          & 66.8          & 58.6          & -           \\
% \textbf{LLaVA-7B +RP}    & \textbf{79.3}          & {\underline{62.8}}    & {\underline{86.4}}    & \underline{66.1}          & \underline{59.8}          & \textbf{33.2}          & \underline{68.2}          & \underline{61.7}          & $+3.06\%$      \\
% \textbf{LLaVA-7B +RP +RM} & \underline{79.2}          & \textbf{62.9} & \textbf{86.8} & \textbf{67.9}          & \textbf{61.3} & \underline{33.1}          & \textbf{68.8}          & \textbf{61.9}          & $+3.91\%$      \\ \bottomrule
% \end{tabular}%
% }
% \end{table*}

% \usepackage{booktabs}
% \usepackage{graphicx}
\begin{table}[h]
\centering
\caption{Performance comparison on 8 MLLM benchmarks (including ``object removal'' data).}
\label{tab:rm_mllm_benchmark}
\resizebox{\columnwidth}{!}{%
\begin{tabular}{@{}cccccc@{}}
\toprule
Model                     & VQA$^{v2}$     & GQA            & POPE           & MMB                              & MMB$^{CN}$     \\ \midrule
LLaVA-7B                  & 78.5           & 62.0           & 85.9           & 64.3                             & 58.3           \\
\textbf{LLaVA-7B + RP}     & \textbf{79.3}  & \underline{62.8} & \underline{86.4} & \underline{66.1}                   & \underline{59.8} \\
\textbf{LLaVA-7B + RP + RM} & \underline{79.2} & \textbf{62.9}  & \textbf{86.8}  & \textbf{67.9}                    & \textbf{61.3}  \\ \bottomrule
                          &                &                &                &                                  &                \\ \toprule
Model                     & MM-Vet         & SQA$^{I}$      & SEED           & $\triangle$ & MMVP           \\ \midrule
LLaVA-7B                  & 30.5           & 66.8           & 58.6           & \multicolumn{1}{c|}{-}           & 24.0           \\
\textbf{LLaVA-7B + RP}     & \textbf{33.2}  & \underline{68.2} & \underline{61.7} & \multicolumn{1}{c|}{$+3.06\%$}   & \underline{27.3}           \\
\textbf{LLaVA-7B + RP + RM} & \underline{33.1} & \textbf{68.8}  & \textbf{61.9}  & \multicolumn{1}{c|}{$+3.91\%$}   & \textbf{28.7}           \\ \bottomrule
\end{tabular}%
}
\end{table}
\begin{table}[h]
\centering
\caption{Results on image difference benchmarks (including ``object removal'' data).}
\label{tab:rm-imgdiff-benchmark}
\resizebox{\columnwidth}{!}{%
\begin{tabular}{@{}ccccc@{}}
\toprule
\multirow{2}{*}{Model}        & \multicolumn{4}{c}{Spot-the-Diff}                                                         \\ \cmidrule(l){2-5} 
                              & BLEU                 & METEOR               & CIDEr-D              & ROUGE-L              \\ \midrule
LLaVA-1.5-7B                  & 8.5                  & 12.0                 & 38.3                 & 30.1                 \\
\textbf{LLaVA-1.5-7B +RP}     & \underline{9.7}      & \textbf{13.0}        & \underline{43.2}     & \underline{30.8}     \\
\textbf{LLaVA-1.5-7B +RP +RM} & \textbf{9.8}         & \textbf{13.0}        & \textbf{46.5}        & \textbf{31.5}        \\ \bottomrule
\multicolumn{1}{l}{}                               & \multicolumn{1}{l}{} & \multicolumn{1}{l}{} & \multicolumn{1}{l}{} & \multicolumn{1}{l}{} \\ \toprule
\multirow{2}{*}{Model}        & \multicolumn{4}{c}{Image-Edit-Request}                                                    \\ \cmidrule(l){2-5} 
                              & BLEU                 & METEOR               & CIDEr-D              & ROUGE-L              \\ \midrule
LLaVA-1.5-7B                  & 15.1                 & 17.8                 & 60.6                 & 45.2                 \\
\textbf{LLaVA-1.5-7B +RP}     & \underline{16.2}     & \textbf{19.5}        & \underline{60.9}     & \textbf{46.7}        \\
\textbf{LLaVA-1.5-7B +RP +RM} & \textbf{16.8}        & \underline{18.6}     & \textbf{63.9}        & \underline{45.7}     \\ \bottomrule
\end{tabular}%
}
\end{table}

\subsection{Evaluation}
We merge the ``object removal'' data with the ``object replacement'' data, making our dataset focus on both object changes and object presence. To test the performance changes of LLaVA-1.5-7B after adding ``object removal'' data, we incorporate this combined data into the original visual instruction tuning dataset of the MLLM and conduct fine-tuning. Then, we evaluate the fine-tuned model on image difference benchmarks and eight MLLM benchmarks, similar to what is presented on the main page.

In the tables, ``RM'' represents ``object removal'' data.

\subsubsection{Results on MLLM Benchmarks} \Cref{tab:rm_mllm_benchmark} shows the performance of LLaVA-1.5-7B finetuned with additional ``object removal'' data on commonly used MLLM benchmarks. With the assistance of ``object removal'' data, LLaVA-1.5-7B achieves further improvements across various benchmarks compared to the model that only uses ``object replacement'' data, with an average increase of 3.91\%.

\subsubsection{Results on Image Difference Benchmarks}
\Cref{tab:rm-imgdiff-benchmark} shows the performance of LLaVA-1.5-7B fine-tuned with our ``object removal'' data on image difference benchmarks. With ``object removal'' data, LLaVA-1.5-7B shows further improvements in its performance on both the MMVP benchmark and the Spot-the-Diff benchmark, surpassing the results achieved with ``object replacement'' data alone. Besides, its scores fluctuate on the Image-Edit-Request benchmark.

\subsubsection{Further Analysis}
The results indicate that the ``object removal'' data has a comprehensive positive impact on LLaVA-1.5-7B, leading to performance improvements in both MLLM benchmarks and image difference benchmarks. However, during our analysis of sample quality, we notice that some of the generated ``object removal'' samples exhibit subpar image quality, with certain image pairs showing inadequate object removal effects. In light of this, employing a more robust inpainting model or applying additional filters to enhance the quality of these image pairs could further optimize the performance of this dataset.

\onecolumn

\section{Examples}
\label{appendix_example}

\begin{figure}[h]
\begin{center}
\centerline{\includegraphics[width=\textwidth]{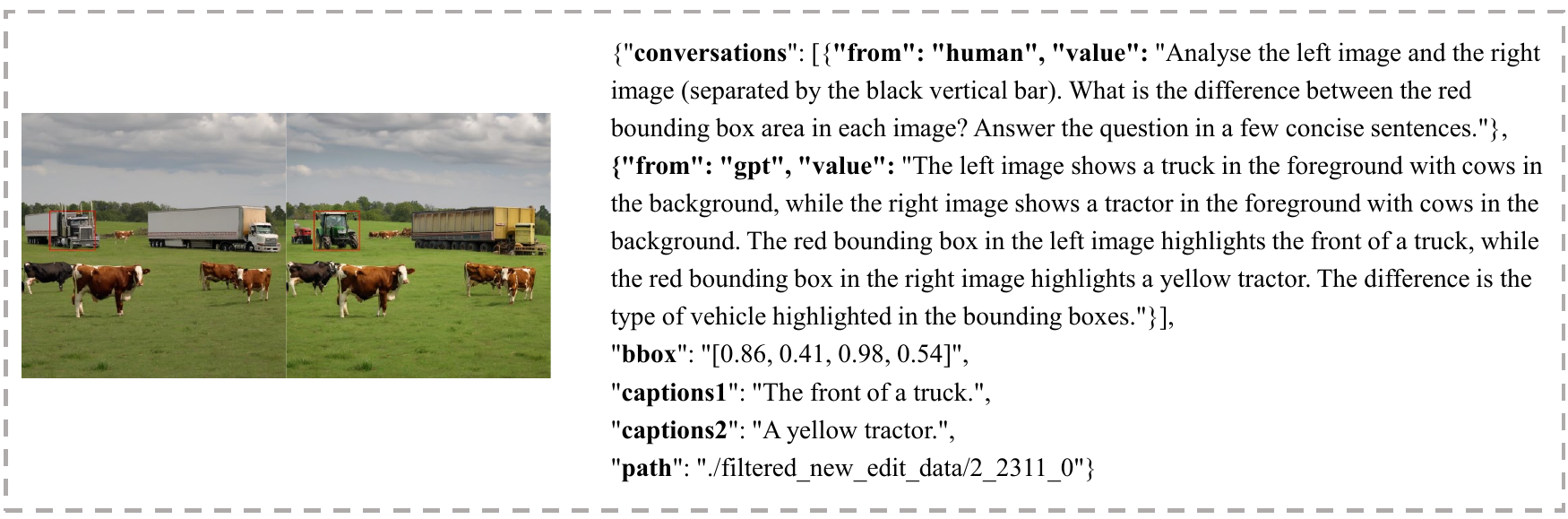}}
\centerline{\includegraphics[width=\textwidth]{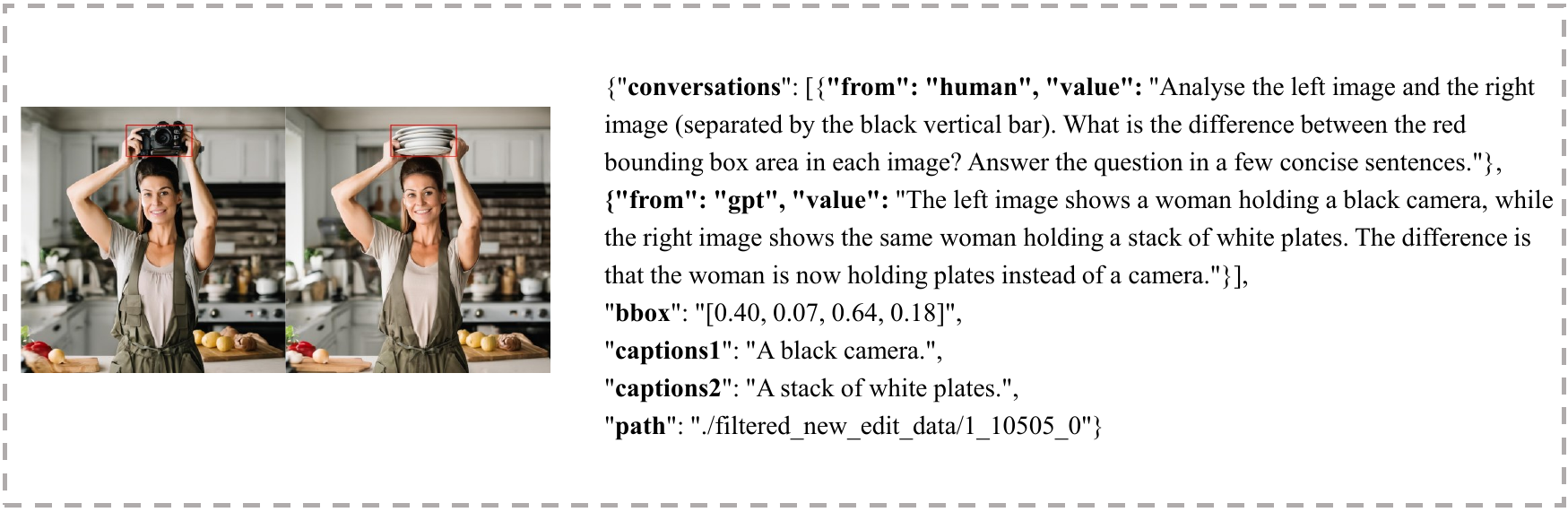}}
\caption{Examples of ``object replacement'' data, including the image pair and the text content in JSON format.}
\label{fig:samples_examples_rp}
\end{center}
\end{figure}

\begin{figure}[h]
\begin{center}
\centerline{\includegraphics[width=\textwidth]{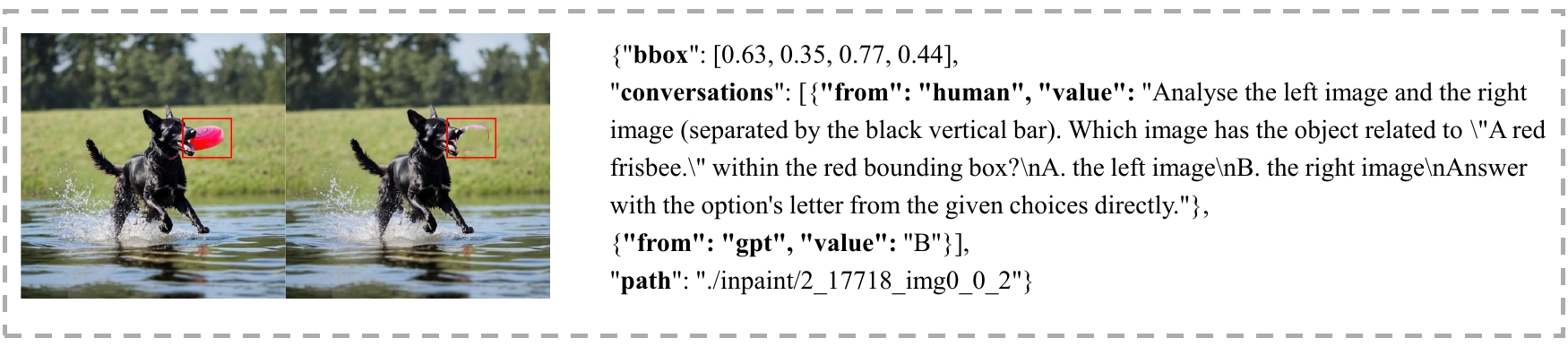}}
\caption{An example of ``object removal'' data, including the image pair and the text content in JSON format.}
\label{fig:samples_examples_rm}
\end{center}
\end{figure}

\twocolumn

\end{document}